%% file: main.tex
\documentclass[runningheads]{llncs}

\usepackage{eccv}

\usepackage{eccvabbrv}

\usepackage{graphicx}
\usepackage{booktabs}

\usepackage[accsupp]{axessibility}  

\usepackage{hyperref}

\usepackage{orcidlink}
\input{preamble}

\begin{document}

\title{Concept-as-Tree: A Controllable Synthetic Data Framework Makes Stronger Personalized VLMs}

\titlerunning{Concept-as-Tree: A Controllable Synthetic Data Framework}

\author{Ruichuan An\inst{1*}
\and
Kai Zeng\inst{1*}
\and
Ming Lu\inst{1}
\and
Sihan Yang\inst{1}
\and \\
Renrui Zhang\inst{2}
\and
Huitong Ji\inst{1}
\and
Hao Liang\inst{1}
\and
Wentao Zhang\inst{1,3,4\dagger}
}

\authorrunning{R.~An et al.}

\institute{$^1$Peking University, $^2$CUHK MMLab, $^3$Zhongguancun Academy, \\
$^4$Beijing Key Laboratory of Data Intelligence and Security (Peking University)
\email{arctanxarc@gmail.com}}

\maketitle

\renewcommand{\thefootnote}{\fnsymbol{footnote}} 
\footnotetext[1]{Equal Contribution.} 
\footnotetext[4]{Corresponding Author.}

\input{sec/0_abstract}
\input{sec/1_intro}
\input{sec/2_related_work}
\input{sec/3_method}
\input{sec/4_exp}
\input{sec/5_conclusion}

\section*{Acknowledgements}
This work is supported by Fundamental and Interdisciplinary Disciplines Breakthrough Plan of the Ministry of Education of China (JYB2025XDXM113), National Natural Science Foundation of China (92470121, 62402016), National Key R\&D Program of China (2024YFA1014003), Zhongguancun Academy (C20250204, C20250602),  Beijing Major Science and Technology Project (Z251100008125043, Z251100008425023), and High-performance Computing Platform of Peking University.

\input{sec/X_suppl}

%
%
\bibliographystyle{splncs04}
\bibliography{main}
\end{document}

%% file: preamble.tex
\usepackage{tabularx}    
\usepackage{makecell}    
\usepackage{tcolorbox}
\usepackage{multirow}

\usepackage[table]{xcolor}
\definecolor{deepgreen}{RGB}{34, 139, 34}
\definecolor{verylightgray}{HTML}{F8F8F8}
\definecolor{mylightgray}{HTML}{EEEEEE}
\definecolor{mygray}{HTML}{CCCCCC}

\usepackage{amsmath}
\usepackage{booktabs}
\usepackage{wrapfig}

%% file: sec/0_abstract.tex
\begin{abstract}
Vision-Language Models (VLMs) have demonstrated exceptional performance in various multi-modal tasks. Recently, there has been an increasing interest in improving the personalization capabilities of VLMs. To better integrate user-provided concepts into VLMs, many methods use positive and negative samples to fine-tune these models. However, the scarcity of user-provided positive samples and the low quality of retrieved negative samples pose challenges for existing techniques. To reveal the relationship between sample and model performance, we systematically investigate the amount and diversity impact of positive and negative samples (easy and hard) on VLM personalization tasks. Based on the detailed analysis, we introduce Concept-as-Tree (CaT), which represents a concept as a tree structure, thereby enabling the data generation of positive and negative samples with varying difficulty and diversity, and can be easily extended to multi-concept scenarios. With a well-designed data filtering strategy, our CaT framework can ensure the quality of generated data, constituting a powerful pipeline. We perform thorough experiments with various VLM personalization baselines to assess the effectiveness of the pipeline, alleviating the lack of positive samples and the low quality of negative samples. Our results demonstrate that CaT equipped with the proposed data filter significantly enhances the capabilities of VLMs across personalization benchmarks. To the best of our knowledge, this work is the first controllable synthetic data pipeline for VLM personalization. 
The code is released at \href{https://github.com/zengkaiya/CaT}{https://github.com/zengkaiya/CaT}.

\keywords{Vision-Language Model \and Personalization \and Synthetic Data}
\end{abstract}

%% file: sec/1_intro.tex
\section{Introduction}
\label{sec:intro}

Vision-Language Models (VLMs)~\cite{bai2023qwenvl, lin2023sphinx,li2023blip} have produced impressive results across various tasks, showcasing their potential as AI assistants~\cite{dorka2024training, zhang2024vision, hong2025dialoguelanguagemodellargescale,zheng2026pearl,xu2025jarvis}. 
Despite their success, VLMs still struggle to generate personalized responses, such as including user-specific identifiers like $\langle \text{Bob} \rangle$ or $\langle \text{Lina} \rangle$ in the conversation.
Several recent studies~\cite{alaluf2025myvlm, nguyenyo, an2024mcllavamulticonceptpersonalizedvisionlanguage, hao2024rememberretrievegenerateunderstanding} have explored the personalization of VLMs to address this challenge and seamlessly incorporate VLMs into everyone's daily life.
Among these, the test-time fine-tuning paradigm exemplified by Yo'LLaVA~\cite{nguyenyo} demonstrates effectiveness. Its training set includes positive, easy-negative, and hard-negative samples, with detailed examples provided in the appendix.
\begin{figure}[t]
    \centering
    \includegraphics[width=0.75\textwidth]{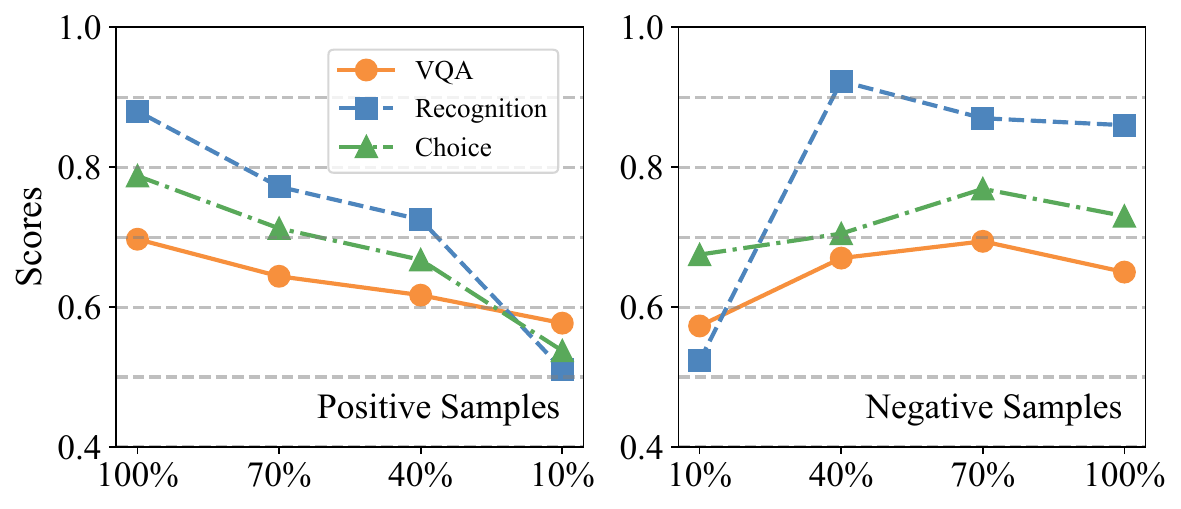}
    \caption{\textbf{The performance of Yo'LLaVA on Yo'LLaVA Dataset.} (Left) Various tasks show a noticeable decrease in performance when the number of positive samples is limited. (Right) As the number of retrieved negative samples increases, the performance of various tasks does not improve consistently. This might be due to uncertainty in the image retrieval process, leading to low-quality negative samples.}
    \label{fig:two_figures}
\end{figure}

Although test-time fine-tuning has shown its effectiveness, its success largely hinges on the availability of both positive and negative samples during the process. In real-world applications~\cite{song2021bob, welch2022leveraging, chawla2013bringing}, positive samples available for personalizing models are often limited.
For instance, users typically provide only 1 to 3 concept images rather than more than 10 for convenience, significantly limiting the model's personalization performance.
Furthermore, the acquisition of negative samples often relies on image retrieval~\cite{meghwani2024enhancing,nguyenyo}, a unpredictable process that complicates the assurance of their quality.
Preliminary experiments, as illustrated in Fig.~\ref{fig:two_figures}, corroborate the limitations. Given these challenges, a natural idea is to use synthetic data to provide controllable supplementation for both positive and negative samples. While this approach may help mitigate the aforementioned issues, three significant challenges remain to be addressed:
\begin{enumerate}
    \item The roles for positive and negative samples in VLM personalization are still not well understood. For instance, is the use of easy negative samples always necessary?
    \item The effect of sample diversity on personalization tasks requires more in-depth investigation. Furthermore, it is important to clarify how the diversity of negative samples affects the performance of the model.
    \item How can we ensure the quality of generated samples? This is closely linked to the model's performance.
\end{enumerate}

\begin{figure*}[t]
    \centering
\includegraphics[width=0.95\textwidth]{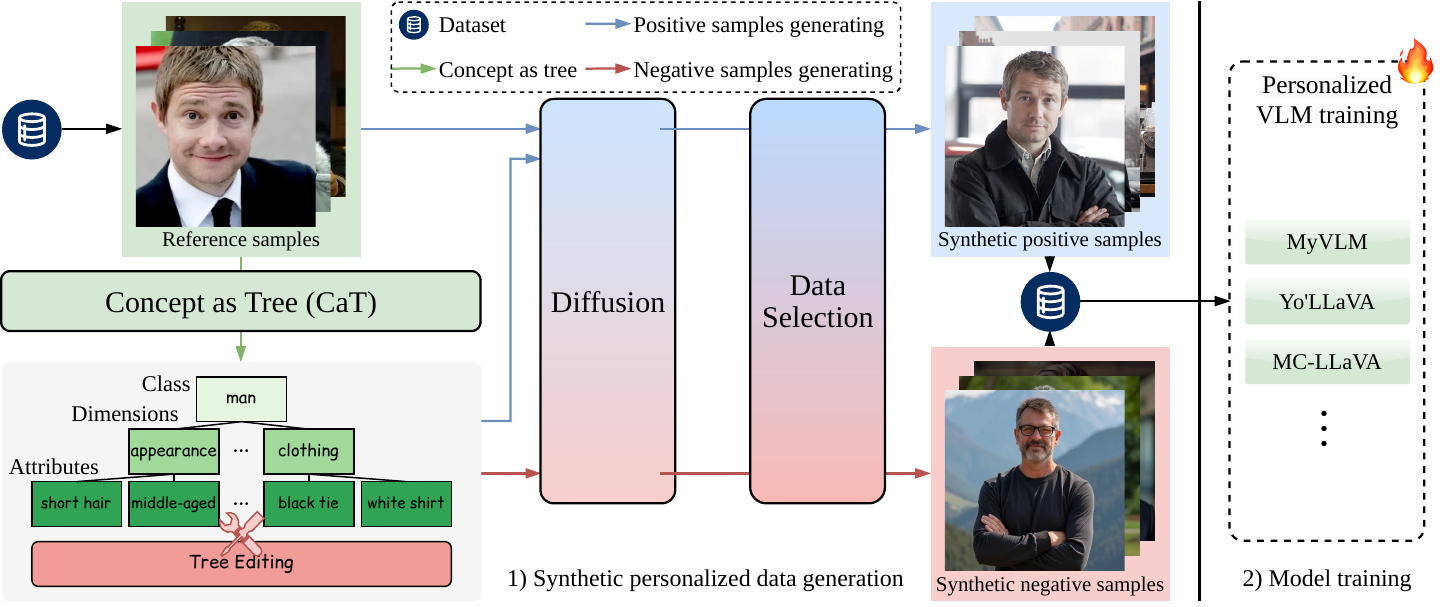}
    \caption{\textbf{Overview of unified and controllable data synthesis pipeline and personalized model training.} After a systematic analysis of positive and negative samples, we utilize LLM and VLM to construct the concept tree and edit it to achieve controllable generation. We then propose a well-designed data selection module and a new metric named PCS score to ensure the quality of synthetic data. The ultimate high-quality data can be used to enhance test-time fine-tuning methods.}
    \label{fig:fig2}
\end{figure*}
To address the challenges mentioned above, we first conducted a series of observational experiments. Our findings revealed that positive samples generally enhance model performance, easy negative samples improve recognition and hard negative samples boost visual question answering capabilities.
We further analyzed easy and hard negative samples with varying diversity and identified specific data diversity requirements for each sample type.
These experiments are essential because they provide significant insights into understanding the roles of different data types and their diversity requirements, ultimately leading to enhanced model performance across various VLM personalization tasks.

Based on these findings, we propose \textbf{C}oncept \textbf{a}s \textbf{T}ree (CaT), a unified and controllable synthetic data framework for VLM personalization.
Specifically, we define a three-layer tree as a representation of concept information in which the root node represents the concept category and the leaf nodes represent its attributes.
By leveraging the power of LLMs and VLMs, we can automate the construction of trees for each concept.
To synthesize positive samples, we directly employ the root node as a prompt for diffusion model. Easy negative samples are generated by altering the root node information and constructing a tree for controllable data generation. Hard negative samples are produced according to concept tree whose leaf nodes are strategically adjusted. 
Moreover, as the number of concepts increases, we naturally merge the trees into forests according to the specific scenarios~\cite{an2024mc} (e.g., two concepts).
Because of the designed tree structure, our framework supports all types of personalized data.
Meanwhile, controlling diversity is equivalent to varying the number and types of editing operations performed on the tree in our framework.

\begin{figure*}[!ht]
    \centering
    \includegraphics[width=0.97\textwidth]{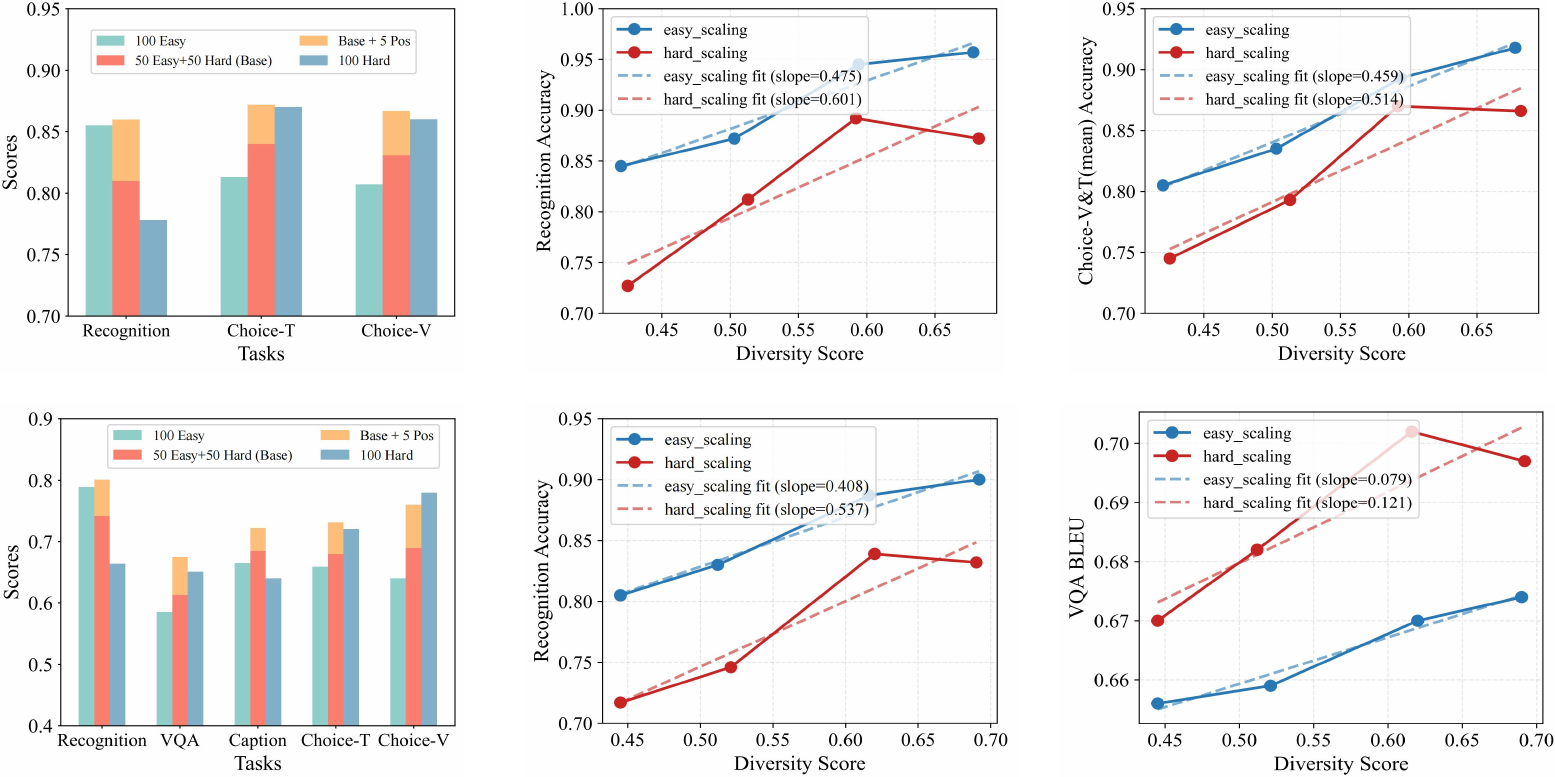}
    \caption{\textbf{Explore the role of positive and negative samples in personalization and their demand for diversity.} (Left) We evaluate the effect of halving positive and negative samples. Positive samples generally improve performance, while easy and hard negatives show task-specific benefits. (Middle \& Right) With fixed sample counts, we vary the diversity of easy and hard negatives. Hard negatives are more sensitive to diversity, and excessive diversity can degrade performance. Diversity scores are computed by clustering retrieved negatives via K-means and measuring distances to cluster centroids. The first line shows results on the Yo'LLaVA dataset; the second line illustrates results on the MC-LLaVa dataset.}
    \label{fig:three_figures}
\end{figure*}

A comprehensive data synthesis process must ensure the quality of generated data. Therefore, we propose a well-designed and easily implementable method for data filtering.
We state that the information in an image includes two types: concept-specific features that are unique to each concept and concept-agnostic features that possess general characteristics~\cite{liu2023vida}.
Therefore, we apply some perturbation to the synthesized images and calculate the distance between the synthesized images and the user-provided images both before and after the perturbation. The difference between these two distances can be defined as a Perturbation-based Concept-Specific (PCS) score, which serves as a key metric for assessing sample quality. 
We set filtering thresholds based on the PCS score to ensure high data quality when filtering various types of generated samples.
Through the aforementioned methods, we achieved a unified and controllable pipeline of data synthesis, as shown in Fig.~\ref{fig:fig2}.


To summarize, our work contributes in multiple aspects:
\begin{itemize}
    \item We systematically study the impact of positive and negative samples and their diversity on VLM personalization.
    \item We propose a comprehensive synthetic data pipeline,  which consists of CaT and a data filtering strategy.
    \item We utilize generated data to conduct extensive experiments on the MyVLM, Yo'LLaVA and MC-LLaVA datasets,  where equipped with the pipeline, all fine-tuning-based methods show significant improvements in various VLM personalization tasks, achieving state-of-the-art results across all datasets.
\end{itemize}

%% file: sec/2_related_work.tex
\section{Related work}
\label{sec:formatting}

\noindent \textbf{Personalization for Vision Language Models.}
Vision-Language Models (VLMs)~\cite{liu2023visual, li2024llava, li2024llava_next} have exhibited remarkable capabilities across diverse domains, such as data mining~\cite{luo2024llm}, fine-grained understanding~\cite{lin2024draw}, and visual question answering~\cite{cao2025move}. 
To seamlessly integrate these models into daily life, there has been growing interest in VLM personalization~\cite{yang2025small,feng2026m2a, an2024mc, hao2025rap, an2025unictokens}.
This task is first introduced by MyVLM~\cite{alaluf2025myvlm}, employing an additional module strategy for improved injection of concept information into VLMs.
YoLLaVA~\cite{nguyenyo} and MC-LLaVA~\cite{an2024mc} utilized efficient end-to-end fine-tuning methods to tackle challenges in both single and multi-concept scenarios.
The performance of these fine-tuning methods is highly dependent on the training samples.
However, users often employ a limited number of concept images (i.e., 1 to 3), and the challenge of acquiring negative samples complicates fine-tuning approaches.
In this study, we investigate the impact of positive and negative samples and their diversity and propose a controllable data synthesis pipeline, thus overcoming challenges. 

\noindent \textbf{Learning from Synthetic Data.}
The use of synthetic data~\cite{jordon2022synthetic, long2024llms, guo2024generative} has been studied across various computer vision tasks, such as classification~\cite{shipard2023diversity}, segmentation~\cite{yang2023freemask}, and object detection~\cite{feng2024instagen}. 
With the rapid consumption of existing data resources driven by the development of Large Language Models (LLMs) and VLMs, researchers aim for models to achieve further improvements using synthetic data~\cite{peng2024synthesize, li2024synthetic, liu2024synthvlm, huang2024modifyingdataaddressgraph}.
Synthetic corpora are generated for instruction fine-tuning of large models. At the same time, large models leverage recaptioned datasets to enhance quality and improve CLIP~\cite{fan2023improving}.
To better support the learning of VLMs, direct generation of text-image pairs is pursued~\cite{liang2024synth}. 
However, existing generation methods only consider class information~\cite{wang2024attributed}. 
These methods rely on simplistic prompt templates, attributes, and class information, either manually defined or LLM-generated~\cite{dunlap2023diversify, wang2024template}.
To address the issue of insufficient data in VLM personalization scenarios, we supplement the dataset by synthesizing concept-centric positive and negative sample data.
We adopt a unified framework to control the generated samples.

\noindent \textbf{Data Quality and Selection.}
The development of LLMs and VLMs relies heavily on data~\cite{dong2025scalable, goyal2024scaling, zeng2025rethinking}. 
The well-known principle is ``garbage in, garbage out." 
High-quality data can significantly enhance model performance~\cite{ridzuan2024review}. 
To ensure data quality, data selection and filtering are common and indispensable practices~\cite{yi2025bridge, chen2024your}.
Primarily, there are rule-based~\cite{awotunde2021intrusion} and model-based methods~\cite{fabris2025efficient}. 
Methods based on LLMs are widely used in data selection. 
In multimodal scenarios, Clip-filter is one of the most commonly used methods and VLMs are also employed as data filters for self-selection and improvement~\cite{wang2024finetuned, hong2024s,zeng2026steervte}.
However, since the $\langle \text{sks} \rangle$ we need to evaluate is not in the vocabulary of the language model, retraining the VLM for each new word is cumbersome.
Thus, we employ a simple yet effective method by perturbing synthetic images. 
We evaluate the concept information contained in the original image by calculating the similarity between the image and the reference image before and after perturbation, and apply a predetermined threshold for selection and filtering.

%% file: sec/3_method.tex
\section{Method}
We present a comprehensive data synthesis pipeline to address the challenge of personalizing Vision-Language Models (VLM). 
This pipeline views the concept information as a tree structure and controls the generation of positive and negative samples via tree operations. 
It also encompasses a well-designed filtering strategy to select the generated samples.
The overall design of our pipeline is illustrated in Fig.~\ref{fig:fig2}. 
Specifically, in part 1, we systematically study the impact of positive and negative samples on VLM personalization tasks. 
Building on this, we introduce the CaT framework and related tree operations in part 2. Furthermore, to ensure the quality of the synthetic samples, we propose a filtering strategy based on the PCS score in part 3.

\begin{figure*}[!ht]
    \centering
\includegraphics[width=0.95\textwidth]{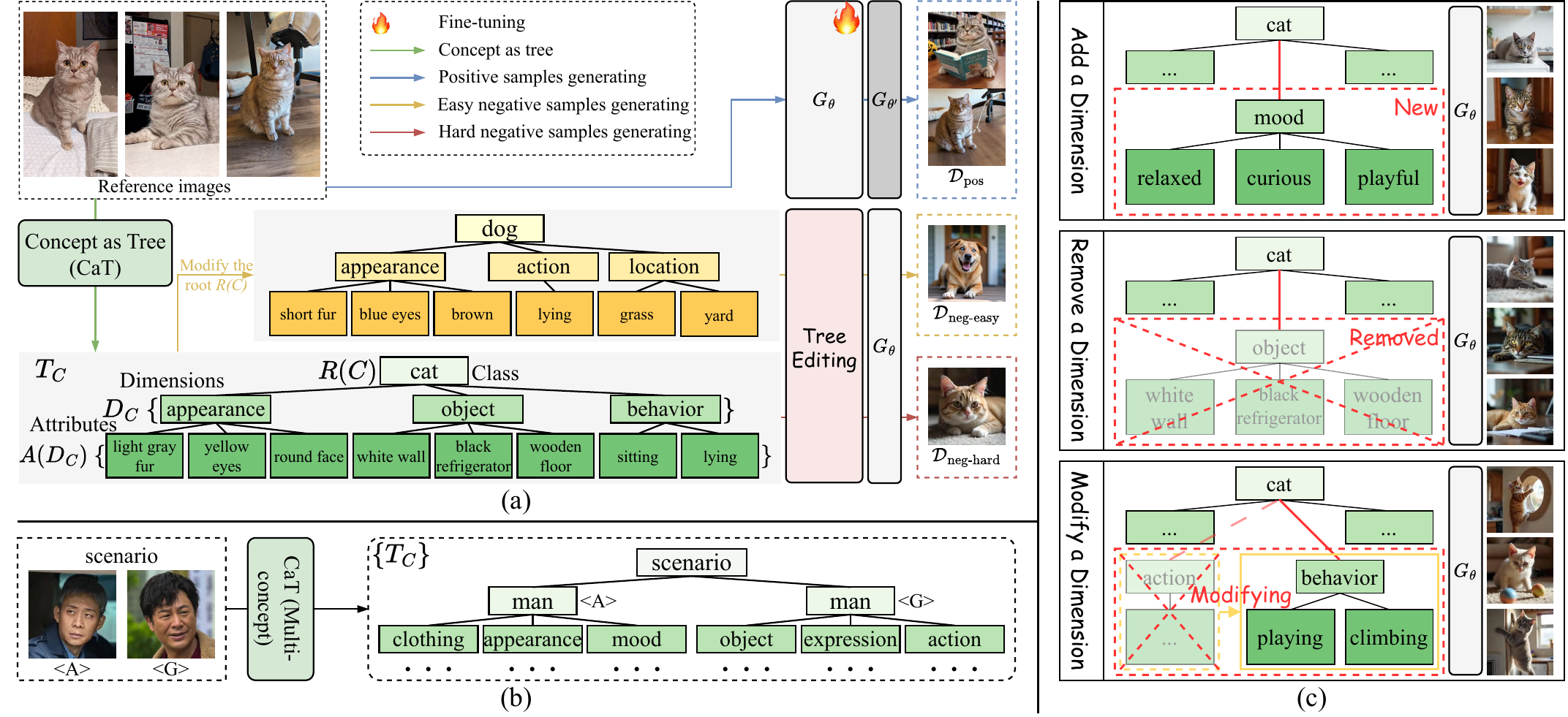}
    \caption{\textbf{Concept tree synthesis and editing operations for sample generation.} (a) A concept tree $T_C$ is constructed using the CaT framework and reference images. Positive samples are synthesized with a fine-tuned diffusion model guided by the root node $R(C)$. Easy negatives are generated by changing the class of $R(C)$, while diverse hard negatives are created by applying three editing operations to the original tree, using an unfine-tuned model. (b) Multiple concept trees can also be merged to support multi-concept personalization data generation. (c) The three operations are visualized: adding a ``mood" dimension alters emotions; removing the ``object" simplifies the scene; modifying a dimension changes behavior.}
    \label{fig:method}
\end{figure*}

\subsection{Impact of Positive and Negative Samples}
\label{sec:impact}
   
In this section, we examine the effects of positive and negative samples respectively and their diversity on the personalization of VLM.
Based on the positive samples provided by existing datasets and negative samples retrieved online, we conducted a series of experiments with the Yo'LLaVA method.
As illustrated in Fig.~\ref{fig:three_figures} left, we found that an increase in positive samples consistently improves outcomes across all tasks. 
While easy negative samples primarily enhance recognition abilities, hard negative samples improve conversational capabilities.
Interestingly, the captioning task fundamentally reflects conversational abilities. 
However, since the evaluation method assesses the presence of a concept identifier related to recognition capabilities, the training data that includes both easy and hard negative samples yields the best performance.
Furthermore, we find that easy negative samples are essential. 
Using only hard negative samples can help the model perform well in conversational tasks, but recognition ability significantly declines. 
However, recognition tasks are the first step for VLMs acting as personal assistants, representing an essential capability that is crucial.
Hence, easy negative samples are irreplaceable due to their diversity and broad sourcing, which offers valuable context for the model's learning process. 

Based on the above mentioned observation, we conducted additional experiments that examined different levels of diversity for negative samples.
Here, diversity is calculated by clustering a set of negative samples and calculating the distance from the cluster center to each sample.
From Fig.~\ref{fig:three_figures} middle and right, keeping the number of positive samples unchanged, it is evident that the gain curves for easy and hard negative samples differ across various tasks, indicating that their diversity requirements are distinct. 
Low diversity of both easy and hard negative samples can simultaneously restrict the model performance trained on them.
When the diversity of hard negative samples is excessively high, it often introduces noise during training, resulting in a decline in model performance.
These experiments clarify the role of diversity and motivate us to manage diversity when creating negative samples.
Therefore, we propose representing the concept as tree to facilitate the generation of negative samples with various difficulties and diversities.

\subsection{Concept as Tree Framework}
\label{3.2}

To integrate new concepts into a pre-trained VLM using synthetic data, we propose the Concept-as-Tree (CaT) framework, which transforms a concept into a structured tree for controllable data generation. Given a user-provided dataset $\mathcal{D}_{\text{user}}$, which contains the positive image samples $I_{\text{pos}} \in \mathcal{D}_{\text{user}}$, our goal is to construct a fine-tuning dataset:
\begin{equation}
    \mathcal{D}_{\text{train}} = \mathcal{D}_{\text{user}} \cup \mathcal{D}_{\text{pos}} \cup \mathcal{D}_{\text{neg-easy}} \cup
    \mathcal{D}_{\text{neg-hard}}
\end{equation}
where $\mathcal{D}_{\text{pos}}$ are synthetic positive samples generated based on $I_{\text{pos}}$, while $\mathcal{D}_{\text{neg-easy}}$ and $\mathcal{D}_{\text{neg-hard}}$ are negative samples obtained via concept tree modifications.

\noindent\textbf{Concept Tree Representation and Construction.} 
To enable controllable data generation, inspired by SSDLLM~\cite{luo2024llm}, we represent each concept $C$ as a hierarchical tree $T_C = (R(C), D_C, A(D_C))$. 
As shown in Fig.~\ref{fig:method}(a), $R(C)$ serves as the root node, encapsulating the high-level category of the concept \textit{(e.g., ``cat'', ``dog'', etc.)}, 
while $D_C = \{D_1, D_2, \dots, D_m\}$ defines a set of attribute dimensions that distinguish various aspects \textit{(e.g., ``appearance'', ``behavior'', ``location'', etc.)}.
Each dimension $D_i$ is associated with a set of attributes $A(D_i)$, collectively forming $A(D_C) = \{A(D_1), A(D_2), \dots, A(D_m)\}$ \textit{(for instance, the ``behavior'' dimension might include attributes like ``sitting'', ``lying'' and ``climbing'', etc.)}.

To automatically construct $T_C$, we adopt a three-step process. 
First, a pre-trained VLM generates textual descriptions from $I_{\text{pos}}$, capturing key visual features (Image Description). 
Next, we aggregate these descriptions at the batch level to extract meaningful dimensions $D_C$ and corresponding attributes $A(D_C)$ (Batch Summarization). 
Finally, to improve accuracy and interpretability, a self-refine mechanism iteratively adjusts $T_C$ based on feedback from a multi-round voting mechanism (Self-Refine). For a given concept image, if it cannot be assigned to the same attribute multiple times, this suggests redundancy or unsuitable attributes. Adjustments enhance the orthogonality and completeness of the attributes.
Detailed prompts and visualizations of the concept tree can be found in the Appendix.

\noindent\textbf{Personalized Data Generation.} After constructing $T_C$, we proceed to generate synthetic samples (see Fig.~\ref{fig:method}(a)). We define an image generation model $G_{\theta}$ for controllable data synthesis and a transformation function $F(R(C), D_C)$ to systematically modify either the tree root or its dimensions.
\begin{itemize}
    \item \textbf{Positive Samples:} Generated by fine-tuning $G_{\theta}$ on user-provided images $I_{\text{pos}}$, conditioned on $R(C)$:
    \begin{equation}
        \mathcal{D}_{\text{pos}} = G_{\theta}(I_{\text{pos}}, R(C))
    \end{equation}

    \item \textbf{Negative Samples:} Constructed by editing $T_C$ to introduce controlled variations.
    \begin{itemize}
        \item \textbf{Easy Negative Samples:} Generated by replacing $R(C)$ with another category $R(C')$ and editing dimensions to $D_C'$, forming a modified tree $T_{C'} = F(R(C'), D_C')$. Synthetic images are produced as $\mathcal{D}_{\text{neg-easy}} = G_{\theta}(T_{C'})$.

        \item \textbf{Hard Negative Samples:} Retaining $R(C)$ while modifying dimensions of $T_C$ to form $T_C' = F(R(C), D_C')$. The generated samples resemble $C$ visually but differ semantically, represented as $\mathcal{D}_{\text{neg-hard}} = G_{\theta}(T_C')$.
    \end{itemize}
\end{itemize}
\noindent \textbf{Multi-Concept Scenarios.} Our CaT framework seamlessly supports multi-concept scenarios by merging different concept trees according to the specific scenario, as illustrated in the Fig.~\ref{fig:method}(b). We generate training data for multiple concepts using the same approach as that used for single concepts. The scalability of the tree structure allows our framework to adapt flexibly to the continuous addition of new concepts, which is a significant advantage of our CaT framework.

\noindent\textbf{Tree Editing Operations.} We define three fundamental tree editing operations $F(R(C), D_C')$ (see Fig.~\ref{fig:method}(c)), which modify $T_C$ to systematically generate negative samples:
\begin{footnotesize}
\begin{equation}
D_C' =
\begin{cases}
    D_C \cup \{D_{\text{new}}\}, & \text{(Add a Dimension)} \\
    D_C \setminus \{D_{\text{remove}}\}, & \text{(Remove a Dimension)} \\
    (D_C \setminus \{D_{\text{old}}\}) \cup \{D_{\text{new}}\}, & \text{(Modify a Dimension)}
\end{cases}
\end{equation}
\end{footnotesize}
where $D_C'$ is the updated dimension set, $D_{\text{new}}$ is an added dimension, and $D_{\text{remove}}$ is a removed one. The modification operation replaces $D_{\text{old}}$ with $D_{\text{new}}$.
The different forms of perturbations and their application frequencies significantly affect the diversity of the generated samples, which will be discussed in the ablation experiments.

\begin{figure*}[ht]
    \centering
\includegraphics[width=0.95\textwidth]{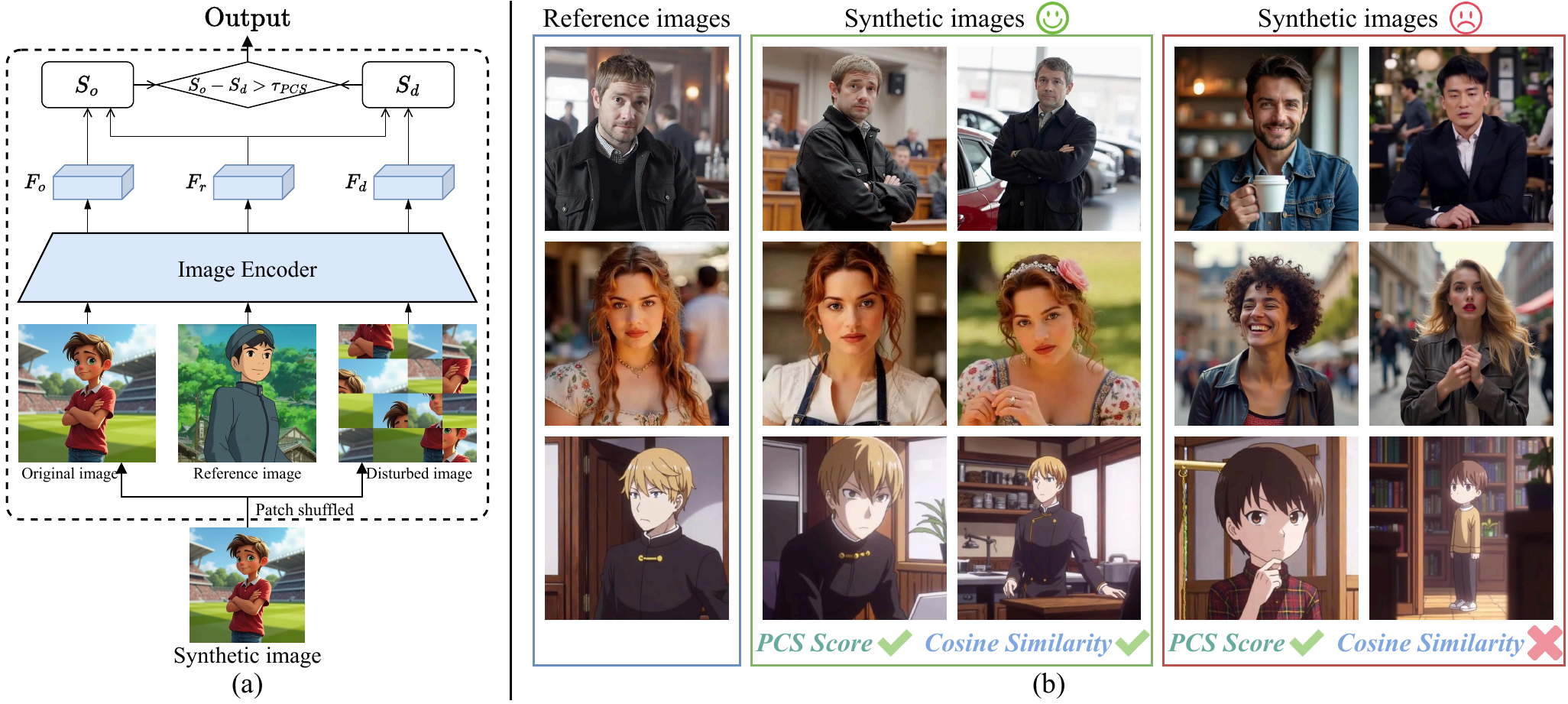}
    \caption{\textbf{PCS-based filtering and visualization of high-quality image selection.} (a) We apply patch shuffling to synthetic images and extract features $F_o$, $F_d$, and $F_r$ from original, disturbed and reference images. Image similarities $S_o$ and $S_d$ are computed to select images with high PCS scores. (b) Images rich in CS information are in green and those rich in CA information are in red. Cosine similarity can not distinguish between two types, but PCS score effectively filters out unqualified ones.}
    \label{fig:PCS_Visualization}
\end{figure*}

\subsection{Sample Filtering with PCS Score}
\label{3.3}
To ensure the quality of synthesized samples, we implement a filtering process for all generated samples. 
We classify the information contained in images into two types: concept-specific (CS) and concept-agnostic (CA). The former refers to features that are distinctive to the concept, such as appearance, while the latter encompasses other elements present in the image, such as background. 
We consider synthetic data predominantly containing CS information to be regarded as high-quality samples. Conversely, samples with more CA information may be categorized as low-quality samples.

Building upon this, we first mix patches from the reference and original images, subsequently calculating the similarity between the altered images and the reference images using the CLIP model. 
The difference in similarities before and after perturbation is defined as the Perturbation-based Concept-Specific (PCS) score. Typically, the PCS score of an image is expected to be high due to the alteration of patches, thereby disrupting the original Concept-Specific (CS) information. 
This disruption causes CLIP to struggle to discern the conceptual relationship between the synthetic image and the reference image, resulting in a significant decrease in similarity between the altered and reference images. However, if the similarity is based on Conceptual-Agnostic (CA) information rather than CS information, the similarity remains relatively unchanged, leading to a low PCS score.  Therefore, we claim that the PCS score can be used to determine the information components in the image. A high PCS score signifies more CS information, while a low PCS score indicates more CA information. 


As illustrated in Fig.~\ref{fig:PCS_Visualization}(a), let the reference image be $I_{\text{reference}} \in \mathcal{D}_{\text{user}}$, the synthetic image before perturbation be $I_{\text{original}}$ and the image after perturbation be $I_{\text{disturbed}}$. 
We extract the visual features $F_r, F_o$ and $F_d$ of the corresponding images using CLIP~\cite{radford2021learning} image encoder $f_\theta$: $F_I=f_{\theta }(I)$
and then calculate the cosine similarity $S_o$ and $S_d$, respectively. 
The cosine similarity ${S}_{x} \in [-1, 1]$ is calculated as:
\begin{equation}
     {S}_{x}=\frac{{F}_{1} {F}_{2}^{\top}}{\parallel {F}_{1} \parallel \cdot \parallel {F}_{2} \parallel }, \label{eq:consine_similarity}
\end{equation}
The difference between $S_o$ and $S_d$ is the PCS score. Based on this score, we can select high-quality synthetic data, i.e., $G(I) = \{ I \mid \text{PCS}(I) > \tau_{PCS} \}$, where $G(I)$ represents high-quality samples and $\tau_{PCS}$ is the threshold for containing more CS information. 
As shown in Fig.~\ref{fig:PCS_Visualization}(b), when only using the filtering method based on cosine similarity, both the images in the green and red boxes achieve high similarity scores. However, the concept in the red box lacks specific visual features of the reference image's concept, such as hair, facial structure and body shape. Instead, the background and other CA information in these images have a greater impact on the similarity. 
However, through the proposed perturbation-based filtering method, we can effectively identify and filter out images with more CA information using the PCS score, while retaining high-quality images with more CS information for better personalization.

%% file: sec/4_exp.tex
\section{Experiments}
\subsection{Experimental Setup}

\noindent \textbf{Implementation Details.}
Regarding the experiments conducted on all datasets, the quantity of positive and negative samples for all the baselines follows their original settings.
The training data used for Baseline (syn) is generated by CaT. 
Baseline (Real+Syn) is articulated as comprising 1 to 3 original concept images and several synthesized positive samples, ensuring the same amount of the original positive samples. 
Baseline (Real+Syn) (Plus) indicates that an equal number of synthesized positive samples have been added to the previous positive samples. All negative sample quantities in the above experiments are strictly controlled for consistency. 
With regard to the synthesized images, we utilize GPT-4o to generate the instruction text pairs required for their training.
Details of baseline, datasets and training hyperparameters can be found in the Appendix.

\begin{table*}[!ht]
\centering
\tiny
\setlength{\tabcolsep}{0.1mm}  
\scalebox{0.9}{
\begin{tabular}{c||c|c|c|c|c|c|c|c|c|c}
\toprule
\textbf{Dataset} & \multicolumn{5}{c|}{\textbf{MC-LLaVA Evaluation}} & \multicolumn{3}{c|}{\textbf{Yo'LLaVA Evaluation}} & \multicolumn{2}{c}{\textbf{MyVLM Evaluation}} \\
\cmidrule{1-11}
\textbf{Method/Task} & \multicolumn{1}{c|}{\textbf{Rec}} & \multicolumn{1}{c|}{\textbf{Choice-V}} & \multicolumn{1}{c|}{\textbf{Choice-T}} & \multicolumn{1}{c|}{\textbf{VQA}} & \multicolumn{1}{c|}{\textbf{Caption}} &  \multicolumn{1}{c|}{\textbf{Rec}} & \textbf{Choice-V} & \textbf{Choice-T} & \multicolumn{1}{c|}{\textbf{Rec}} & \textbf{Caption} \\

\midrule
\textbf{GPT-4o+Prompt} & 0.746 & 0.888 & 0.712 & 0.728 & 0.836 & 0.856 & 0.932 & 0.897 & 0.891 & 0.969 \\
\midrule

\textbf{MyVLM(Real)} & 0.795 & 0.779 & - & 0.640 & 0.714 & 0.911 & 0.897 & - & 0.938 & 0.921 \\
\textbf{MyVLM(Syn)} & 0.788 & 0.772 & - & 0.653 & 0.711 & 0.908 & 0.895 & - & 0.935 & 0.918 \\
\textbf{MyVLM(Real+Syn)} & 0.799 & 0.783 & - & 0.662 & 0.722 & 0.916 & 0.907 & - & 0.945 & 0.926 \\
\textbf{MyVLM(Real+Syn)(P)} & 0.827 & 0.805 & - & 0.685  & 0.740  & 0.945  & 0.916  & - & 0.960  & 0.947 \\
& \textbf{+3.2\%} & \textbf{+2.6\%} & \textbf{-} & \textbf{+4.5\%} & \textbf{+2.6\%} & \textbf{+3.4\%} & \textbf{+1.9\%} & \textbf{-} & \textbf{+2.2\%} & \textbf{+2.6\%} \\

\midrule
\textbf{Yo'LLaVA(Real)} & 0.841 & 0.801 & 0.703 & 0.643 & 0.701  & 0.924 & 0.929 & 0.883 & 0.964 & 0.931 \\
\textbf{Yo'LLaVA(Syn)} & 0.840 & 0.805 & 0.705 & 0.654 & 0.699 & 0.921 & 0.925 & 0.885 & 0.962 & 0.927 \\
\textbf{Yo'LLaVA(Real+Syn)}  & 0.851 & 0.817 & 0.711 & 0.662 & 0.714 & 0.928 & 0.933 & 0.891 & 0.969 & 0.938 \\
\textbf{Yo'LLaVA(Real+Syn)(P)} & 0.885 & 0.845 & 0.738 & 0.682 & 0.754 & 0.946 & 0.943 & 0.900 & 0.981 & 0.950 \\
& \textbf{+4.4\%} & \textbf{+4.4\%} & \textbf{+3.5\%} & \textbf{+3.9\%} & \textbf{+5.3\%} & \textbf{+2.2\%} & \textbf{+1.4\%} & \textbf{+1.7\%} & \textbf{+1.7\%} & \textbf{+1.9\%} \\

\midrule
\textbf{MC-LLaVA(Real)}  & 0.917 & 0.890 & 0.727 & 0.684 & 0.750 & 0.947 & 0.941 & 0.893 & 0.975 & 0.959 \\
\textbf{MC-LLaVA(Syn)} & 0.920 & 0.892 & 0.731 & 0.695 & 0.755 & 0.951 & 0.945 & 0.896 & 0.978 & 0.963 \\
\textbf{MC-LLaVA(Real+Syn)} & 0.928 & 0.890 & 0.739 & 0.704 & 0.768 & 0.958 & 0.947 & 0.901 & 0.984 & 0.968 \\
\textbf{MC-LLaVA(Real+Syn)(P)} & 0.963 & 0.928 & 0.760 & 0.726 & 0.803 & 0.977 & 0.953 & 0.910 & 0.987 & 0.971 \\
& \textbf{+4.6\%} & \textbf{+3.8\%} & \textbf{+3.3\%} & \textbf{+4.2\%} & \textbf{+5.3\%} & \textbf{+3.0\%} & \textbf{+1.2\%} & \textbf{+1.7\%} & \textbf{+1.2\%} & \textbf{+2.2\%} \\
\midrule
\textbf{RAP-MLLM} & 0.747 & 0.832 & 0.709 & 0.424  & 0.711 & 0.845 & 0.917 & 0.874 & 0.870 & 0.937 \\
\bottomrule
\end{tabular}
}
\caption{\textbf{Results of synthetic data used in different personalization methods.} Synthetic data improves all methods across three datasets. MC-LLaVA, trained on original positive data combined with all synthetic data, achieves performance comparable to GPT-4o.
Choice-T is a text-based task; MyVLM cannot perform it as it requires concept images for embedding loading. 
Numbers after "+" denote the new SOTA's improvement over the original baseline. Evaluation tasks: Real = Original Data, Syn = Synthetic Data, P = Plus. Evaluation metrics: Rec = Recognition Accuracy, Choice-V = Choice-V Accuracy, Choice-T = Choice-T Accuracy, VQA = BLEU, Caption = Recall. \textbf{Higher} values are preferred for all metrics.}
\label{tab:main_results}
\end{table*}

\subsection{Personalized Capability from Synthetic Data}
The ability to recognize concepts is fundamental to personalized VLM. 
As illustrated in Tab.~\ref{tab:main_results}, it is evident that after training with extra synthesized data, the model's recognition ability shows an average improvement of 4.1\% on the MC-LLaVA dataset, 2.9\% on the Yo'LLaVA dataset, and 1.7\% on the MyVLM dataset.
However, training solely on synthesized data does not yield consistent improvements, which may be attributed to a distribution shift between synthesized data and the real data. 
Within the MC-LLaVA framework, despite the use of purely synthesized data, performance still surpasses the baseline, which is attributed to the utilization of visual information relevant to concepts in the MC-LLaVA methods.
Including original real images of concepts in the training data can mitigate this phenomenon, resulting in a modest improvement.

We further evaluate the capability of our method to address questions related to personalized concepts.
The introduction of synthesized data yields an increase of up to 3.8\% in choice-based tasks, 4.2\% in Visual Question Answering (VQA) tasks, and 5.3\% in captioning tasks, achieving competitive performance with GPT-4o in certain scenarios.
Training with purely synthesized data aligns with the trends observed in recognition tasks. 
Notably, on text-related tasks such as Choice-T and VQA, it consistently enhances model performance. 
This improvement stems from our targeted modifications in negative sample generation, where specific features of positive samples are altered. By reducing the model's reliance on these adjusted features, it learns to focus more on the concept's intrinsic attributes. As a result, the model achieves higher accuracy in concept-related multiple-choice tasks and generates more precise descriptions of the concept’s defining characteristics.



Our CaT framework also supports the synthesis of multi-concept personalized training data. By applying attribute constraints across different concept trees within the same scene, we ensure that the synthesized positive samples do not share overlapping attributes. Experimental results in Tab.~\ref{tab:main_results_multi} show that this significantly enhances the capability of multi-concept personalized models.

Overall, the results obtained from training with purely synthesized data are comparable to those from purely real data, demonstrating the quality assurance of our synthesized data.
In scenarios with limited real data, training in conjunction with our synthesized data achieves results that surpass the baseline, preserving the personalization capabilities of VLMs under data scarcity.
The stable improvements stem from our thorough understanding of concept information and the highly controllable CaT framework used during data synthesis.
Moreover, augmenting real data with additional synthesized data significantly surpasses all baselines, reflecting the efficacy of synthesized data. 

\begin{table}[!h]
\centering
\scriptsize
\setlength{\tabcolsep}{1mm}  
\renewcommand{\arraystretch}{0.95}  
\begin{tabular}{c||c|c|c|c|c}
\toprule
\textbf{Dataset} & \multicolumn{5}{c}{\textbf{MC-LLaVA Evaluation (Multi-Concept)}} \\
\cmidrule{1-6}
\textbf{Method/Task} & \textbf{Rec} & \textbf{Choice-V} & \textbf{Choice-T} & \textbf{VQA} & \textbf{Caption} \\
\midrule
\textbf{GPT-4o+Prompt} & 0.822 & 0.889 & 0.680 & 0.651 & 0.816 \\
\midrule
\textbf{Yo'LLaVA(Real)} & 0.729 & 0.602 & 0.594 & 0.557 & 0.611 \\
\textbf{Yo'LLaVA(Syn)} & 0.715 & 0.613 & 0.579 & 0.541 & 0.625 \\
\textbf{Yo'LLaVA(Real+Syn)} & 0.732 & 0.608 & 0.597 & 0.569 & 0.628 \\
\textbf{Yo'LLaVA(Real+Syn)(P)} & 0.760 & 0.634 & 0.618 & 0.580 & 0.647 \\
& \textbf{+3.1\%} & \textbf{+3.2\%} & \textbf{+2.4\%} & \textbf{+2.3\%} & \textbf{+3.6\%} \\

\midrule
\textbf{MC-LLaVA(Real)} & 0.845 & 0.905 & 0.695 & 0.611 & 0.763 \\
\textbf{MC-LLaVA(Syn)} & 0.832 & 0.918 & 0.682 & 0.598 & 0.776 \\
\textbf{MC-LLaVA(Real+Syn)} & 0.849 & 0.911 & 0.698 & 0.624 & 0.780 \\
\textbf{MC-LLaVA(Real+Syn)(P)} & 0.881 & 0.933 & 0.714 & 0.636 & 0.798 \\
& \textbf{+3.6\%} & \textbf{+2.8\%} & \textbf{+1.9\%} & \textbf{+2.5\%} & \textbf{+3.5\%} \\
\midrule
\textbf{RAP-MLLM} & 0.688 & 0.690 & 0.656 & 0.423 & 0.748 \\
\bottomrule
\end{tabular}
\caption{\textbf{Results of synthetic data used in multi concept scenarios.} Our CaT framework also well supports multi-concept scenarios. When trained with a combination of original positive data and all synthetic data, both Yo'LLaVA and MC-LLaVA gain an impressive improvement. The numbers following the ``+" symbol represent the improvement of the new SOTA compared to original baseline. Real = Original Data; Syn = Synthetic Data; P = Plus. \textbf{Higher} values are preferred for all metrics.}
\label{tab:main_results_multi}
\end{table} 

\subsection{Ablations and Analysis}
\noindent \textbf{The Use of PCS Score Metric.} We compare filtering strategy based on the PCS score with the cosine similarity-based filtering strategy that relies on image features. 
The results presented in Tab.~\ref{tab:filter_strategy} show that the performance of synthetic data without any filtering is even worse than the original Yo'LLaVA~\cite{nguyenyo} baseline. This is because the images synthesized by the generative model may contain low-quality samples that are blurry, unrealistic, or unnatural, which can interfere with the model's training process. 
After applying the cosine similarity-based filtering using image features, some low-quality images or those inconsistent with the style of the reference images are filtered out, leading to improved model performance. 
This demonstrates the necessity of a filtering strategy for synthetic data. 
However, relying solely on conventional image similarity filtering may fail to remove images that achieve high similarity scores based solely on background, style, or other CA information. 
Our proposed PCS score-based filtering strategy addresses this limitation by perturbing the synthetic images and calculating the difference in similarity scores before and after perturbation. 
This approach effectively eliminates the interference of CA information, ultimately selecting images with a higher proportion of CS information. This enables further improvement in the model's personalization performance in all tasks.

\begin{table}[!h]
\centering
\setlength{\tabcolsep}{6pt}  
\scriptsize
\begin{tabular}{c|ccccc}
\toprule
 Filter Strategy & \multicolumn{1}{c}{\textbf{Rec}} & \multicolumn{1}{c}{\textbf{Choice-V}} & \multicolumn{1}{c}{\textbf{Choice-T}} & \multicolumn{1}{c}{\textbf{VQA}} & \multicolumn{1}{c}{\textbf{Caption}} \\
\midrule
None & 0.835 & 0.793 & 0.691 & 0.633 & 0.697  \\
 Cosine Similarity & 0.862 & 0.816 & 0.716 & 0.652 & 0.726 \\
PCS Score & \textbf{0.885} & \textbf{0.845} & \textbf{0.738} & \textbf{0.682} & \textbf{0.754} \\
\bottomrule
\end{tabular}
\caption{Ablation of different filtering metrics.}
\label{tab:filter_strategy}
\end{table}

\noindent \textbf{Tree Editing Operations.} We investigate the impact of tree editing operations on hard negative samples within the MC-LLaVA dataset. 
Index \textit{A} represents the Yo'LLaVA baseline without any tree editing operation. The diversity score is defined for a set of data; although there may be a distribution shift between retrieved and generated data, their diversity can still be compared.
In Tab.~\ref{tab:tree}, the comparison between \textit{A}, \textit{B}, \textit{E}, and \textit{G} clearly demonstrates that different tree editing operations lead to variations in the diversity of generated data, resulting in varying task performance.
Both Add and Modify operations can enhance diversity to some extent, with Add yielding a greater increase than Modify. In contrast, removing decreases the diversity of generated results due to a reduction in combinable attributes.
In accordance with Fig.~\ref{fig:three_figures}, high-diversity negative samples lead to a more pronounced improvement in tasks related to conversation.
However, excessive diversity can introduce noise, thus degrading the performance of the model.
For the results of \textit{B}, \textit{C}, \textit{D}, \textit{E}, \textit{F}, \textit{G} and \textit{H}, we observe that an increase in the number of editing operations per instance amplifies their impact on diversity and results. 
Therefore, our CaT framework can provide precise control over this process, ensuring that optimal diversity leads to stable improvements.

\begin{table}[!h]
\centering
\footnotesize 
\setlength{\tabcolsep}{0.3mm} 
\renewcommand{\arraystretch}{0.95}
\begin{tabular}{c|c|c|c|c|c|c|c}
\toprule
\multirow{2}{*}{Methods} & \multirow{2}{*}{Index} & \multicolumn{2}{c|}{TEO} &\multirow{2}{*}{Diversity} & \multicolumn{3}{c}{Dataset: MC-LLaVA}  \\
\cmidrule{3-4}
\cmidrule{6-8}
& & Category & Times &  & Rec & VQA & Caption \\
\midrule
\multirow{8}{*}{\makecell{Yo'LLaVA\\ +CaT}} 
& A & None & 0 & 0.497 & 0.841 & 0.643 & 0.701 \\
& B & Add & 1 & 0.563 & 0.852 & 0.655 & 0.723 \\
& C & Add & 2 & 0.642 & 0.866 & 0.674 & 0.736 \\
& D & Add & 3 & 0.708 & 0.834 & 0.654 & 0.717 \\
& E & Remove & 1 & 0.451 & 0.827 & 0.621 & 0.687 \\
& F & Remove & 2 & 0.379 & 0.801 & 0.605 & 0.663 \\
& G & Modify & 1 & 0.542 & 0.847 & 0.652 & 0.718 \\
& H & Modify & 2 & 0.613 & 0.859 & 0.671 & 0.734 \\
\bottomrule
\end{tabular}
\caption{\textbf{Ablation study on tree editing operations (TEO).} The type and number of operations can both alter the diversity of synthesized data.}
\label{tab:tree}
\end{table}


\noindent \textbf{Data Component.}
We analyze the key components of the generated datasets. 
The original training data consists of provided positive samples and retrieved negative samples. 
Then we examine the role of synthetic data in these three components by first adding synthetic positive samples and then progressively replacing the retrieved negative samples with synthetic negative samples. At each step, we advance incrementally while keeping the other components unchanged.
As shown in Tab.~\ref{tab:data_component}, the original Yo'LLaVA already possesses a certain level of personalization capability. 
However, as samples are progressively transformed into synthesized data, performance on several tasks continues to improve, demonstrating the effectiveness of our CaT approach. The enhancements brought by incorporating easy and hard negative samples vary across different tasks, which is consistent with our observations in observation experiments.

\begin{table}[!ht]
\centering
\setlength{\tabcolsep}{0.4pt}  
\scriptsize
\begin{tabular}{l|ccccc}
\toprule
 Module/Task & \multicolumn{1}{c}{\textbf{Rec}} & \multicolumn{1}{c}{\textbf{Choice-V}} & \multicolumn{1}{c}{\textbf{Choice-T}} & \multicolumn{1}{c}{\textbf{VQA}} & \multicolumn{1}{c}{\textbf{Caption}} \\
\midrule
Yo'LLaVA & 0.841 & 0.801 & 0.703 & 0.643 & 0.701  \\
 w/ Syn Positive & 0.858 \tiny \textbf{(+1.7\%)} & 0.816 \tiny \textbf{(+1.5\%)} & 0.716 \tiny \textbf{(+1.3\%)} & 0.656 \tiny \textbf{(+1.3\%)} & 0.721 \tiny \textbf{(+2.0\%)} \\
w/ Syn Easy Neg & 0.873 \tiny \textbf{(+1.5\%)} & 0.824 \tiny \textbf{(+0.8\%)} & 0.721 \tiny \textbf{(+0.5\%)} & 0.667 \tiny \textbf{(+1.1\%)} & 0.727 \tiny \textbf{(+0.6\%)} \\
 w/ Syn Hard Neg & 0.885 \tiny \textbf{(+1.2\%)} & 0.845 \tiny \textbf{(+2.1\%)} & 0.738 \tiny \textbf{(+1.7\%)} & 0.680 \tiny \textbf{(+1.3\%)} & 0.754 \tiny \textbf{(+2.7\%)} \\
\bottomrule
\end{tabular}
\caption{\textbf{Ablation study on data component.}}
\label{tab:data_component}
\end{table}

\begin{wrapfigure}{r}{0.5\textwidth}
    \centering
    \includegraphics[width=0.46\textwidth]{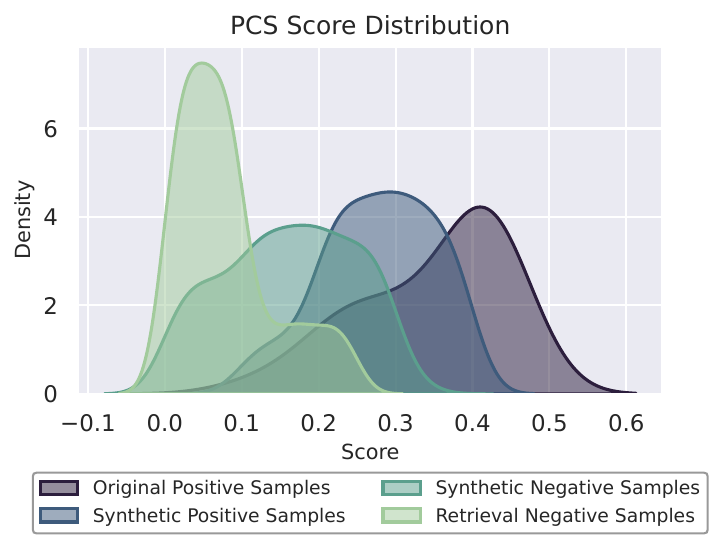}
    \caption{\textbf{The distribution of PCS scores across different datasets.}}
    \label{fig:PCSS_Ablation}
\end{wrapfigure}

\noindent \textbf{Quality of Synthesized Data.}
As discussed in Sec.~\ref{3.3}, a higher PCS score indicates that an image contains more concept-specific information, representing a higher-quality personalized sample. 
From the results shown in Fig.~\ref{fig:PCSS_Ablation}, we observe that the distributions of PCS scores for the original positive samples and the synthetic positive samples are quite similar, demonstrating that our synthetic positive samples can effectively serve the same purpose as the original positive samples. 
Additionally, we find that the proportion of low PCS scores in the original retrieved negative samples is significantly higher than that in our synthetic negative samples, indicating that the quality of synthetic negative samples surpasses that of retrieved negative samples. 

%% file: sec/5_conclusion.tex
\section{Conclusion}
In this work, we present a new roadmap to enhance the personalization capability of Vision-Language Models (VLMs) via leveraging synthetic positive and negative images from advanced generative models. 
Based on our solid and systematic observation experiment, we propose a unified and controllable data synthesis pipeline, which consists of Concept as Tree framework and a data selection module. Remarkably, our pipeline requires only 1 to 3 user-provided images to work, synthesizing high-quality samples that enable VLMs to perform comparably to models trained on real images in data-scarce scenarios. 
This pipeline addresses the critical challenge of insufficient user-provided concept data, offering a controllable, interpretable, and data-efficient solution for VLM personalization tasks, paving the way for VLMs to become more practical human assistants.



%% file: sec/X_suppl.tex


\clearpage
\appendix

\section{More Implementation Details}
\label{sec:exp_setup}
\noindent \textbf{Dataset.}
We utilize datasets from Yo’LLaVA~\cite{nguyenyo} and MyVLM~\cite{alaluf2025myvlm}. Yo’LLaVA consists of 40 categories of objects, buildings and people, with 4 to 10 images available per category for training and validation. In contrast, MyVLM encompasses 29 object categories, each containing 7 to 17 images, with 4 images designated for training and the remainder for validation. Additionally, we incorporate the Single concept portion of MC-LLaVA~\cite{an2024mc} dataset, which is a challenging dataset meticulously constructed from movies, featuring textual descriptions generated with the assistance of GPT-4. MC-LLaVA includes 50 scenarios that encompass various characters and objects, amounting to a total of 118 concepts. 

\noindent \textbf{Model Settings.}
 For all LLMs and VLMs involved in the CaT framework, we utilized GPT-4o~\cite{achiam2023gpt}. 
 After obtaining the concept tree, we synthesize positive and negative samples using the DreamBooth~\cite{ruiz2023dreambooth} fine-tuned FLUX.1-dev model~\cite{flux2024}. The positive samples required for fine-tuning are derived from 1 to 3 original concept images.  
 Finally, we select images using the CLIP ViT-L/14 visual encoder~\cite{radford2021learning} with our proposed filtering method based on the PCS score. 
 Regarding the personalization of VLMs, we test four methods: MyVLM~\cite{alaluf2025myvlm}, Yo'LLaVA~\cite{nguyenyo}, MC-LLaVA~\cite{an2024mc} and RAP-MLLM~\cite{hao2024rememberretrievegenerateunderstanding}. All configurations adhered to the original papers, and we use LLaVA-1.5-13B~\cite{liu2023visual} as the VLM backbone for all experiments.

\noindent \textbf{Hyperparatemers.}
The hyperparameters for fine-tuning the FLUX model can be referenced from DreamBooth~\cite{ruiz2023dreambooth}. 
In the filtering module, the patch size used in patch shuffle is set to 14, consistent with the CLIP visual encoder we employ. Additionally, based on the distribution of PCS scores from the four datasets, we set the PCS score thresholds for synthesized positive and hard negative samples to 0.3 and 0.1, respectively.
For easy negative samples, which do not require CS information, we merely employed conventional text-to-image similarity~\cite{radford2021learning} filtering to ensure that the images matched the prompts.

\section{More Details on the CaT Method}
\noindent \textbf{Comparison with Related Work.}
Our CaT framework adopts the concept representation structure from SSD-LLM~\cite{luo2024llm}. However, since this representation is not directly applicable to customized data synthesis, we refrain from direct comparison. Beyond this, we introduce concept tree editing to enhance tree diversity and mitigate similarity among customized samples. We also propose, for the first time, multi-concept tree synthesis to accommodate scenarios involving multiple customized concepts.
It is worth emphasizing that CaT is not a prompt engineering method, but rather a management framework for constructing diverse prompts efficiently. Through a three-level hierarchy of nodes, dimensions, and attributes, our structure readily yields multi-sample flat attribute sets, structured prompts, or scene-graph-like representations as prompt inputs. Furthermore, our design naturally extends to multi-concept settings without producing overly similar concepts.

\noindent \textbf{Cost of the Entire Synthetic Data Pipeline.}
For one concept: (1) FLUX DreamBooth tuning with 3 reference images: $\sim$240\,s on a single A100; (2) Synthesizing 630 images (10 positive, 100 hard negative, 100 easy negative, with $3\times$ filtering) at 5\,s per image on an A100: $\sim$400\,s with $8\times$A100 parallelism; (3) GPT-4o-based QA generation (300 pairs): $\sim$60\,s. Total: $\sim$\textbf{700\,s} per concept.
Given that 7B LLaVA personalization fine-tuning requires $\sim$\textbf{1~hour} (i.e., 3,600\,s) on 8 A100s, the $\sim$700\,s overhead is considered acceptable.

\section{More Observation Experiments}
\begin{figure}[!h]
    \centering
    \includegraphics[width=0.95\textwidth]{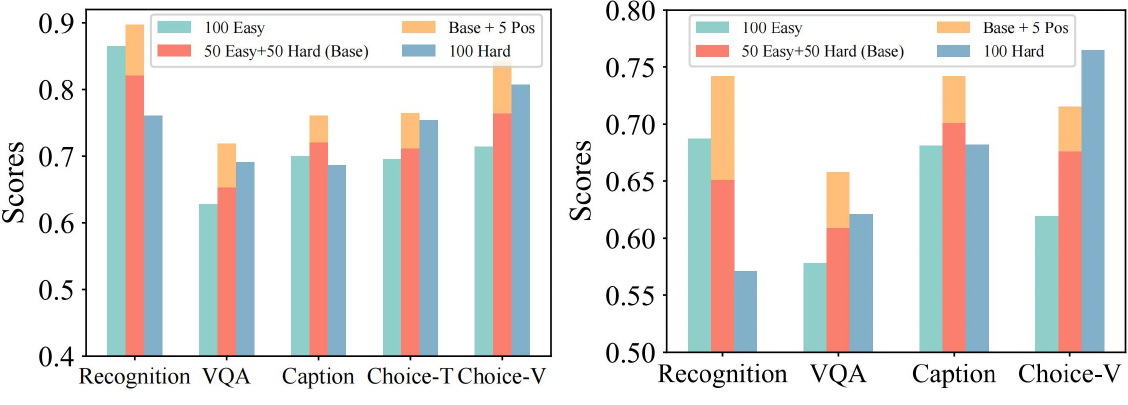}
    \caption{More analysis of positive and negative samples.}  
    \label{fig:more_analysis}
    \vspace{-5mm}
\end{figure}

\noindent \textbf{The Role of Positive and Negative Samples.}
We also use MC-LLaVA and MyVLM model to verify the role of positive and negative samples on the MC-LLaVA dataset. The experimental results are in Fig.~\ref{fig:more_analysis}, which are consistent with the conclusions of the Yo'LLaVA model.

\noindent \textbf{More Observation Experiment on PCS Score and Diversity.}
We retrieved and generated negative samples for each concept in Yo'LLaVA dataset. As demonstrated in Tab.~\ref{tab:pcss_yollava}, our generated hard negative samples demonstrated their high quality in terms of PCS score and diversity measure.

\renewcommand\arraystretch{1.2}
\setlength\tabcolsep{4pt}
\begin{table}[h]
    \scriptsize
    \centering
\begin{tabular}{|c|c|c|c|c|}
\hline
 & $< 0.1$ & $> 0.1$ and $< 0.3$ & $> 0.3$ & diversity \\ \hline
Retrieval Negative Samples & 49.6\% & 49.7\% & 0.7\%  & 0.54 \\ \hline
Synthetic Negative Samples & 35.9\% & 63.2\% & 0.9\%  & 0.68 \\ \hline
\end{tabular}
\vspace{2mm}
\caption{Distribution of PCS score on Yo'LLaVA dataset.}
\label{tab:pcss_yollava}
\end{table}
\vspace{-1mm}

\section{More Ablation Experiments}
\noindent \textbf{Different Image Encoder.}
We conducted additional experiments using new visual encoders and VLMs in the filtering process. As shown in Tab.~\ref{tab:encoder}, performances between individual encoders are comparable, while combined encoders achieve better results. Furthermore, using stronger VLMs leads to additional performance gains, but brings more costs.

\renewcommand\arraystretch{1.2}
\setlength\tabcolsep{6.1pt}
\renewcommand{\arraystretch}{1.0}
\begin{table}[h]
    \scriptsize
    \centering
\begin{tabular}{|c|c|c|c|c|c|}
\hline
Method / Metric & Rec & Choice-V & Choice-T & VQA & Caption  \\ \hline
CLIP & 0.885 & 0.845 & 0.738 & 0.682 & 0.754 \\ \hline
DINOv2 & 0.897 & 0.851 & 0.727 & 0.679 & 0.752 \\ \hline
Combined {\tiny(encoders)} & 0.904 & 0.854  & 0.745 & 0.687 & 0.760 \\ \hline
GPT-4o & 0.906 & 0.862  & 0.752 & 0.695 & 0.765 \\ \hline
\end{tabular}
\vspace*{-4pt}
\vspace{2mm}
\caption{Ablation of different filtering methods, including visual encoders and VLM.}
\label{tab:encoder}
\vspace{-2mm}
\end{table}

\begin{figure}[t]
    \centering
    \includegraphics[width=0.7\textwidth]{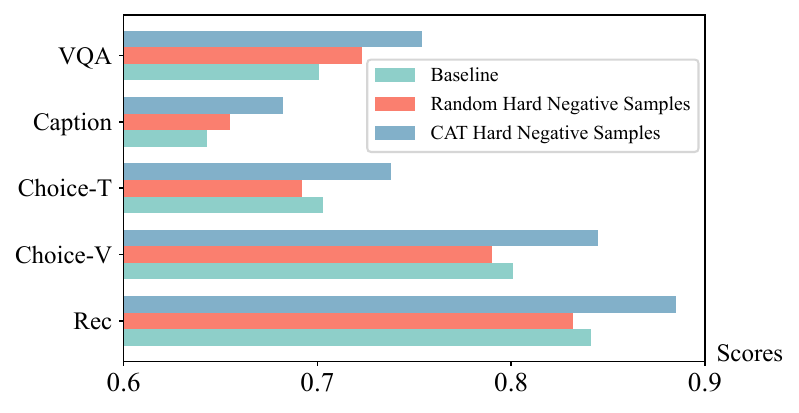}
    \caption{A comparison of hard negative samples.}
    \label{fig:hard_negative_generation}
    \vspace{-5mm}
\end{figure}

\noindent \textbf{Hard Negative Sample Generation.}
We compare different methods for synthesizing hard negative samples, and the results are presented in Fig.~\ref{fig:hard_negative_generation}. Our CaT framework ensures the diversity of synthetic samples, leading to significant improvements across all tasks. In contrast, the random synthesis of hard negative samples lacks control over the diversity of the results. The baseline using these data may overfit to a narrow range of negative samples, leading to an inability to achieve improvements across all tasks.

\section{Visualization of Concept Trees}
\label{sec:cat}
\begin{figure*}[t]
    \centering
    \includegraphics[width=\textwidth]{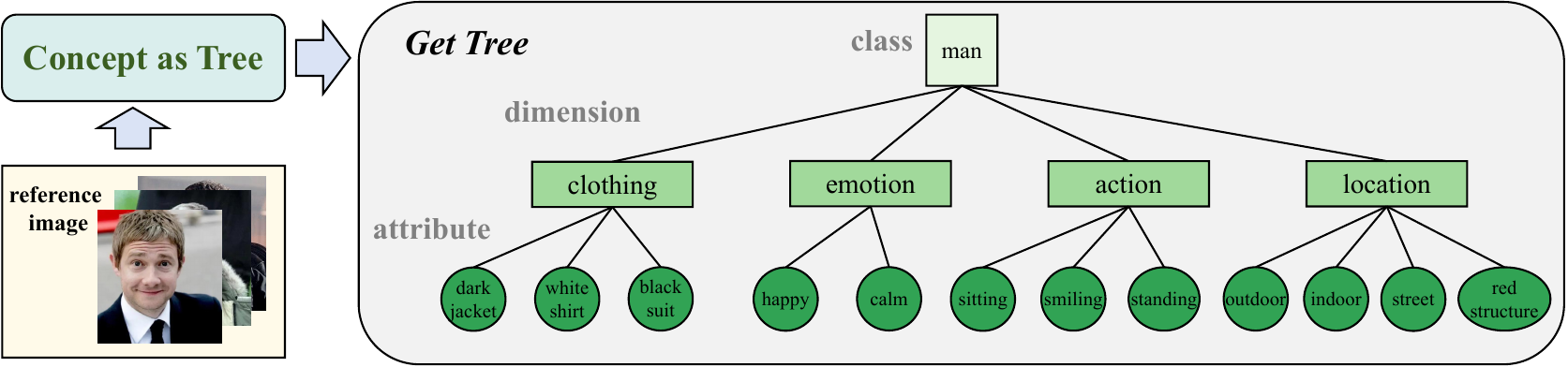}
    \caption{Using the CaT framework, we generate a concept tree for the ``man" concept. The CaT framework can accurately extract the visual information associated with this concept, such as clothing and location.}
    \label{fig:man_get}
\end{figure*}

\begin{figure*}[ht]
    \centering
    \includegraphics[width=\textwidth]{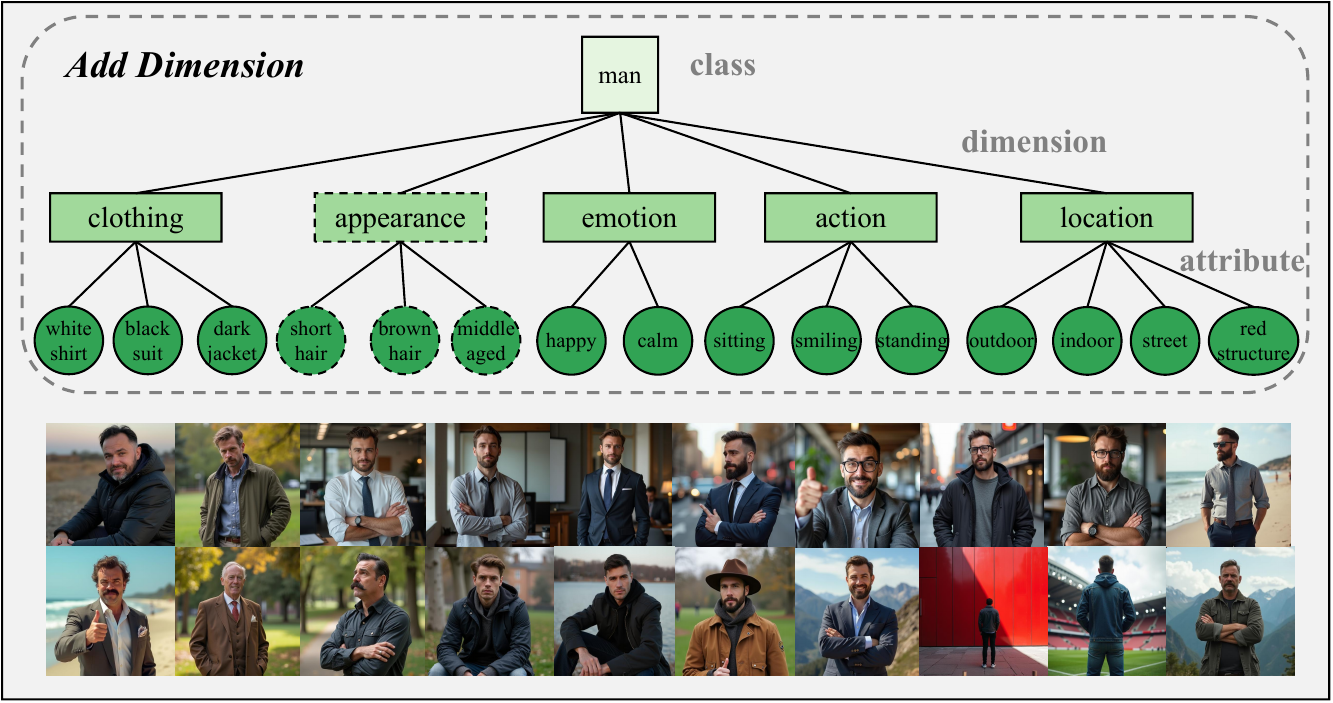}
    \caption{Adding an ``appearance" dimension to the original concept tree results in significant changes in people's appearance, such as hair length and color, which are combined with the original attributes to create a variety of styles of images}
    \label{fig:man_add}
\end{figure*}

\begin{figure*}[ht]
    \centering
    \includegraphics[width=\textwidth]{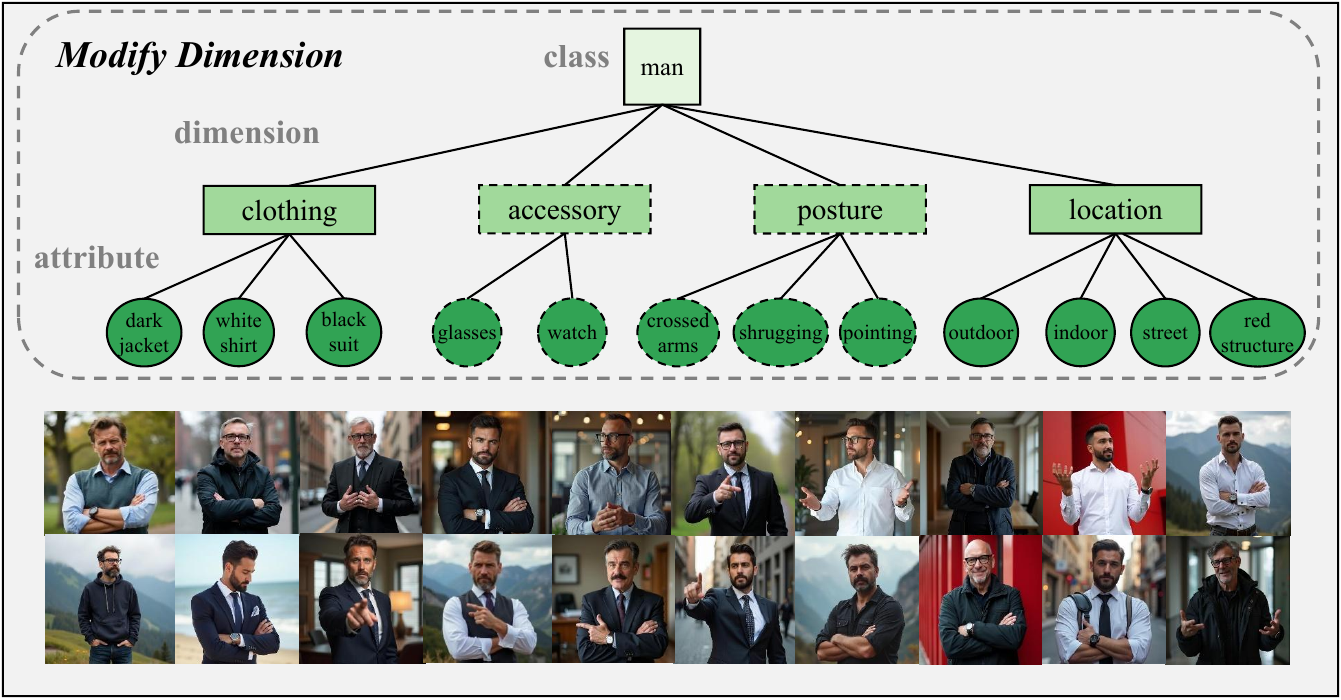}
    \caption{Change the ``emotion" and ``action" dimensions in the original concept tree to the ``accessory" and ``poster" dimensions, allowing for various combinations of male accessories, standing posture, and movements.}
    \label{fig:man_modify}
\end{figure*}

\begin{figure*}[ht]
    \centering
    \includegraphics[width=\textwidth]{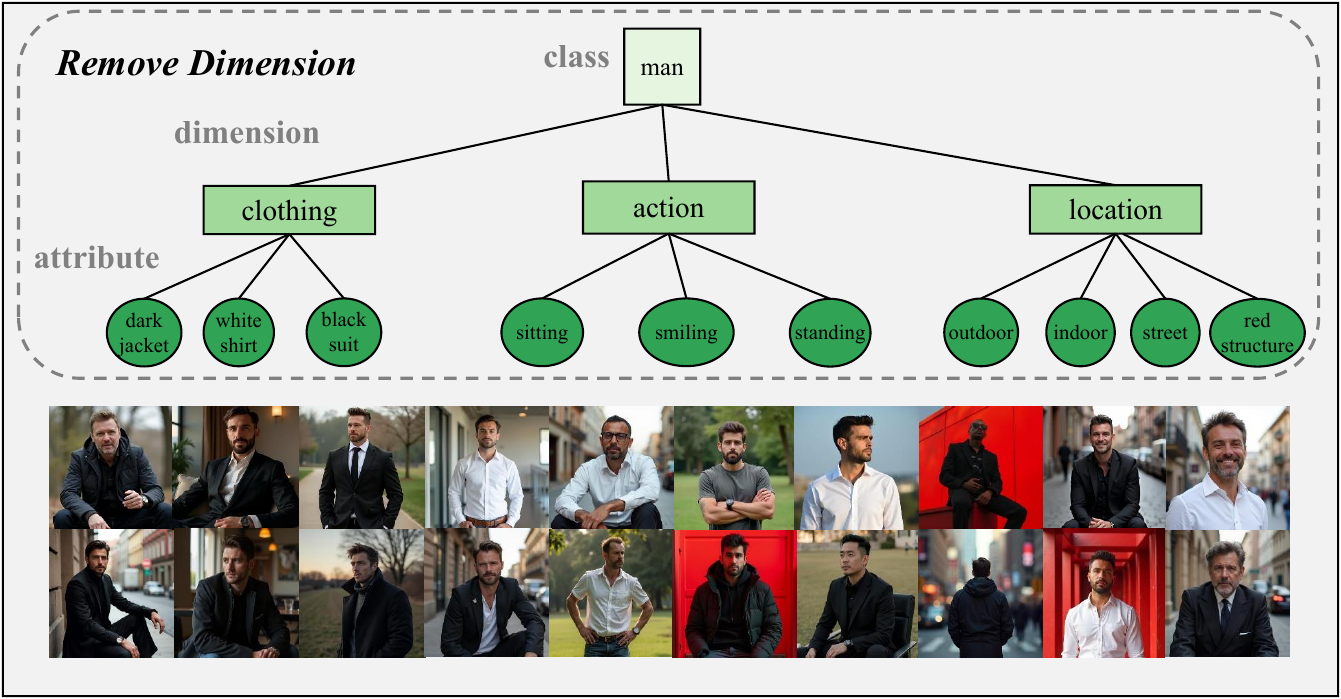}
    \caption{Removing the ``emotion" dimension from the original concept tree, the previously happy, excited, and varied expressions almost uniformly turn into silence, which reduces the diversity of images.}
    \label{fig:man_remove}
\end{figure*}

\begin{figure*}[ht]
    \centering
    \includegraphics[width=\textwidth]{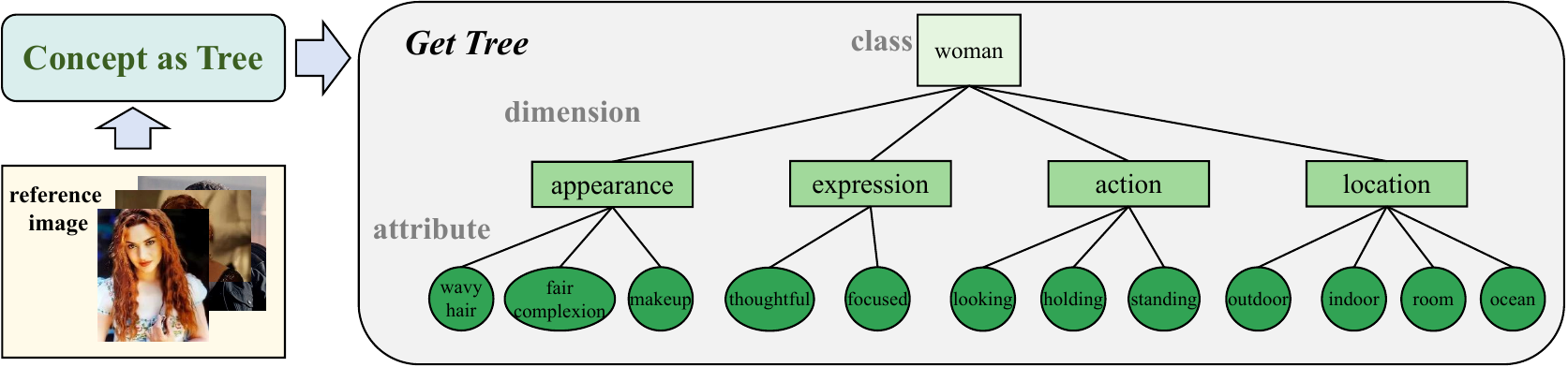}
    \caption{Using the CaT framework, we generate a concept tree for the ``woman" concept. The CaT framework can accurately extract the visual information associated with this concept, such as expression and location.}
    \label{fig:woman_get}
\end{figure*}

\begin{figure*}[ht]
    \centering
    \includegraphics[width=\textwidth]{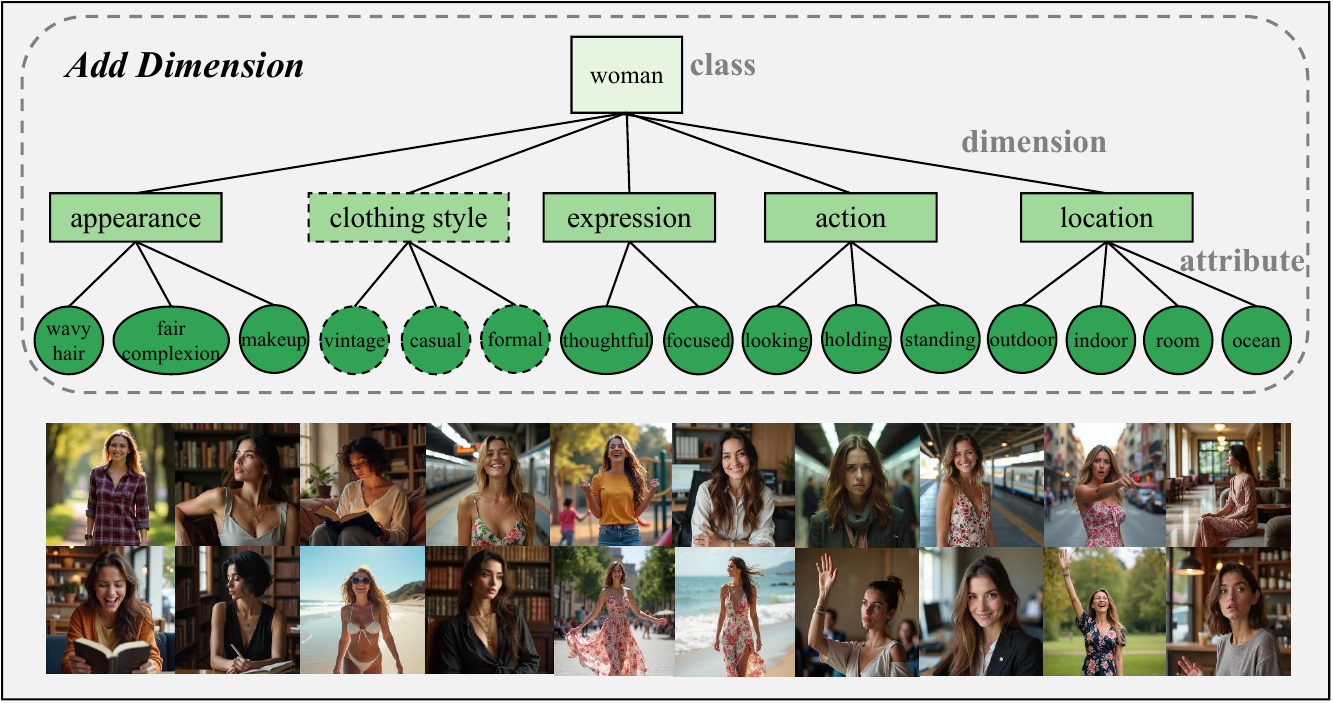}
    \caption{Adding a ``clothing style" dimension to the original concept tree results in a significant change in woman's attire, bringing multiple styles to the images.}
    \label{fig:woman_add}
\end{figure*}

\begin{figure*}[ht]
    \centering
    \includegraphics[width=\textwidth]{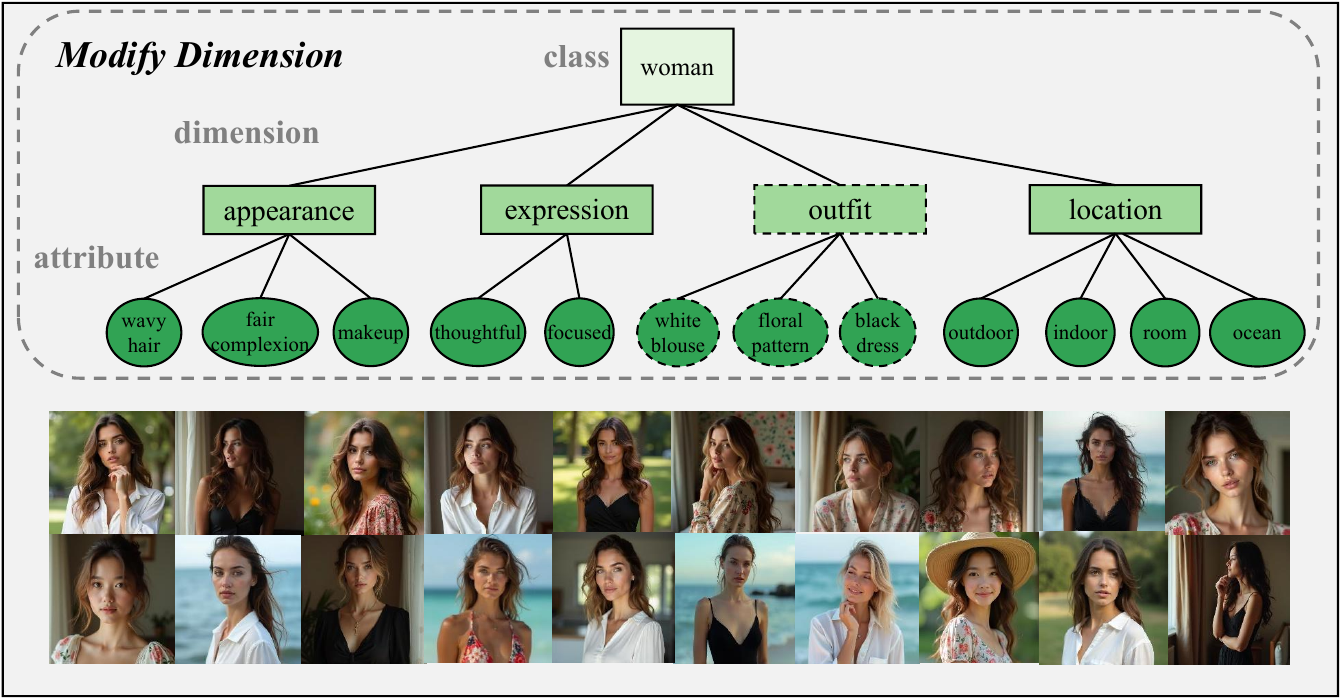}
    \caption{Change the ``action" dimension in the original concept tree to the ``outfit" dimension, so that woman's clothing is no longer just one or two types of reference pictures}
    \label{fig:woman_modify}
\end{figure*}

\begin{figure*}[ht]
    \centering
    \includegraphics[width=\textwidth]{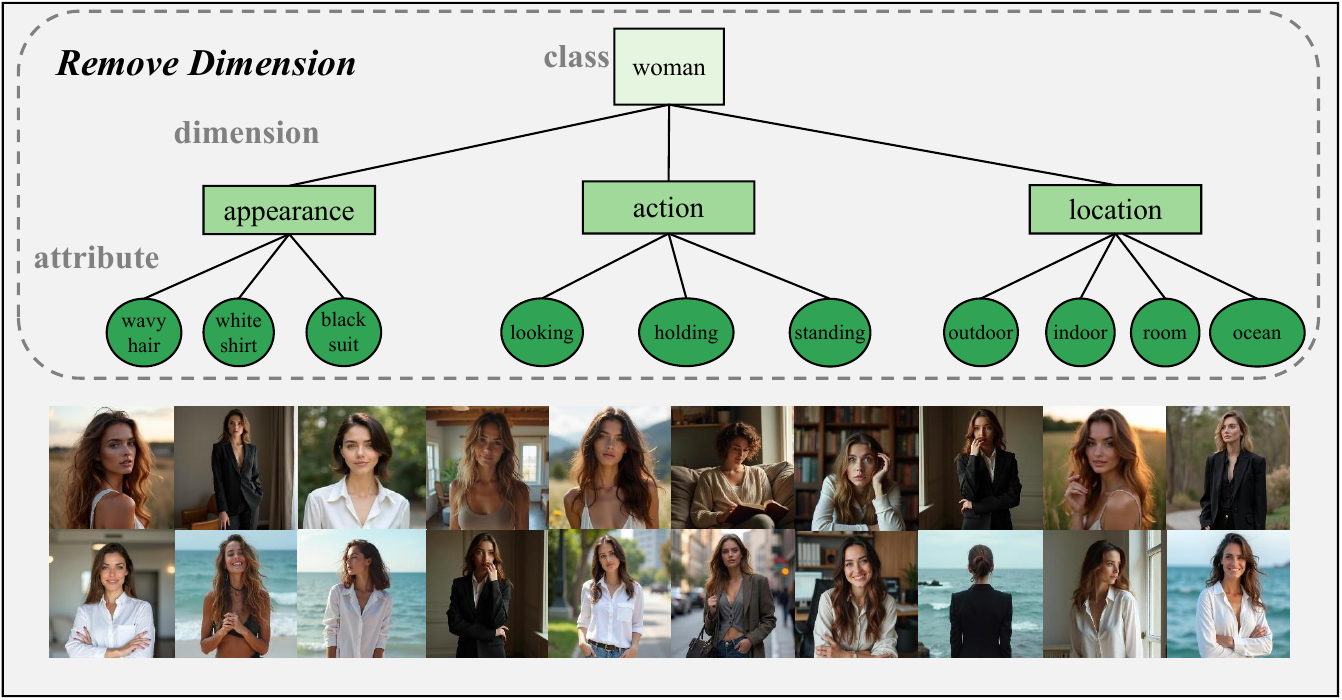}
    \caption{Removing the ``expression" dimension from the original concept tree leaves almost only a calm expression on the woman's face, which reduces the diversity of images}
    \label{fig:woman_remove}
\end{figure*}

\begin{figure*}[ht]
    \centering
    \includegraphics[width=\textwidth]{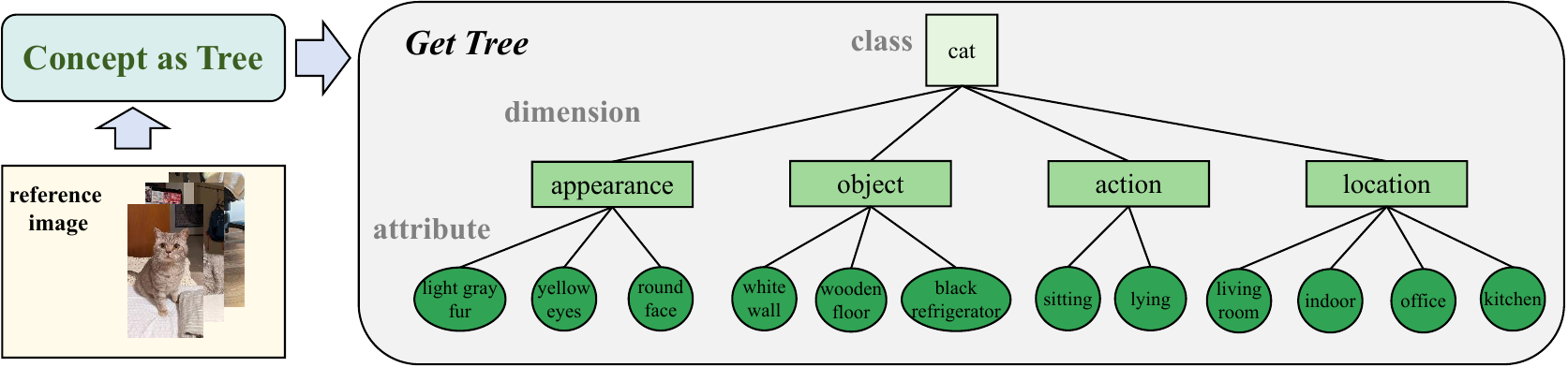}
    \caption{Using the CaT framework, we generate a concept tree for the ``cat" concept. The CaT framework can accurately extract the visual information associated with this concept, such as appearance and action.}
    \label{fig:cat_get}
\end{figure*}

\begin{figure*}[ht]
    \centering
    \includegraphics[width=\textwidth]{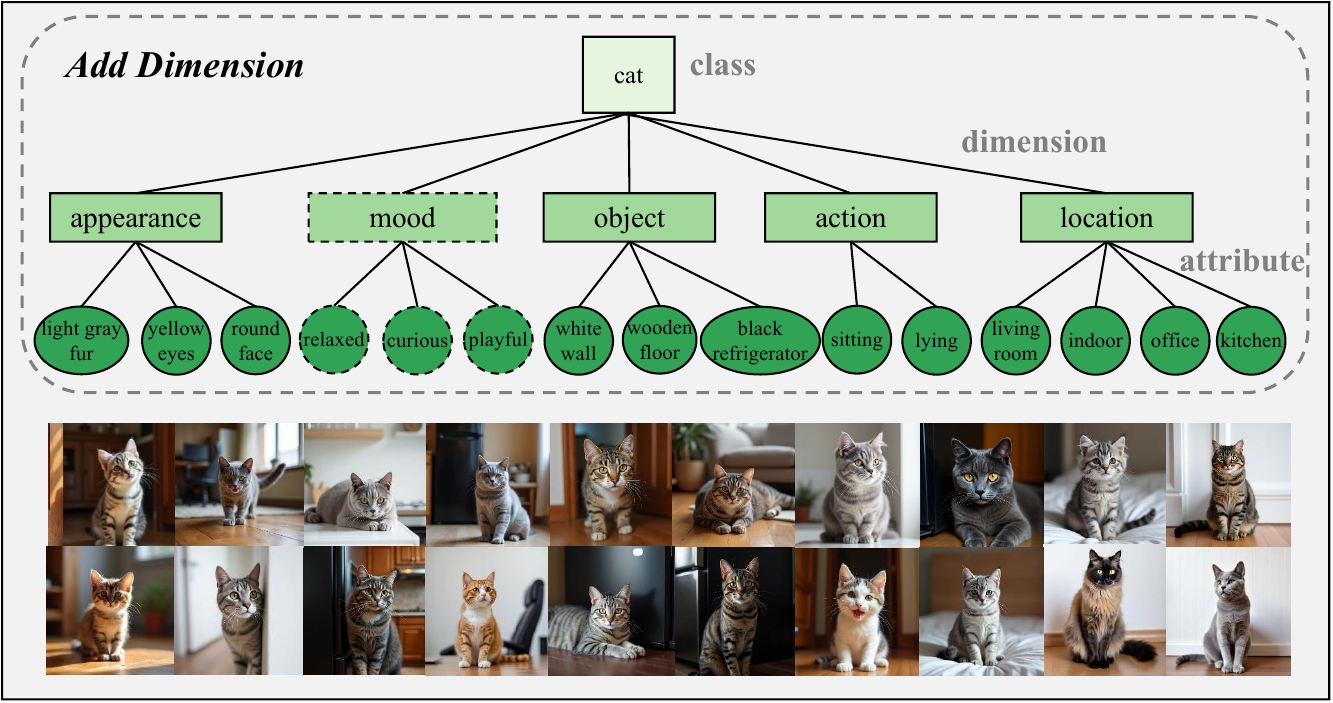}
    \caption{Adding a ``mood" dimension to the original concept tree significantly changes the expression and posture of the cat, making it more interesting compared to the previously silent cat in the reference image.}
    \label{fig:cat_add}
\end{figure*}

\begin{figure*}[ht]
    \centering
    \includegraphics[width=\textwidth]{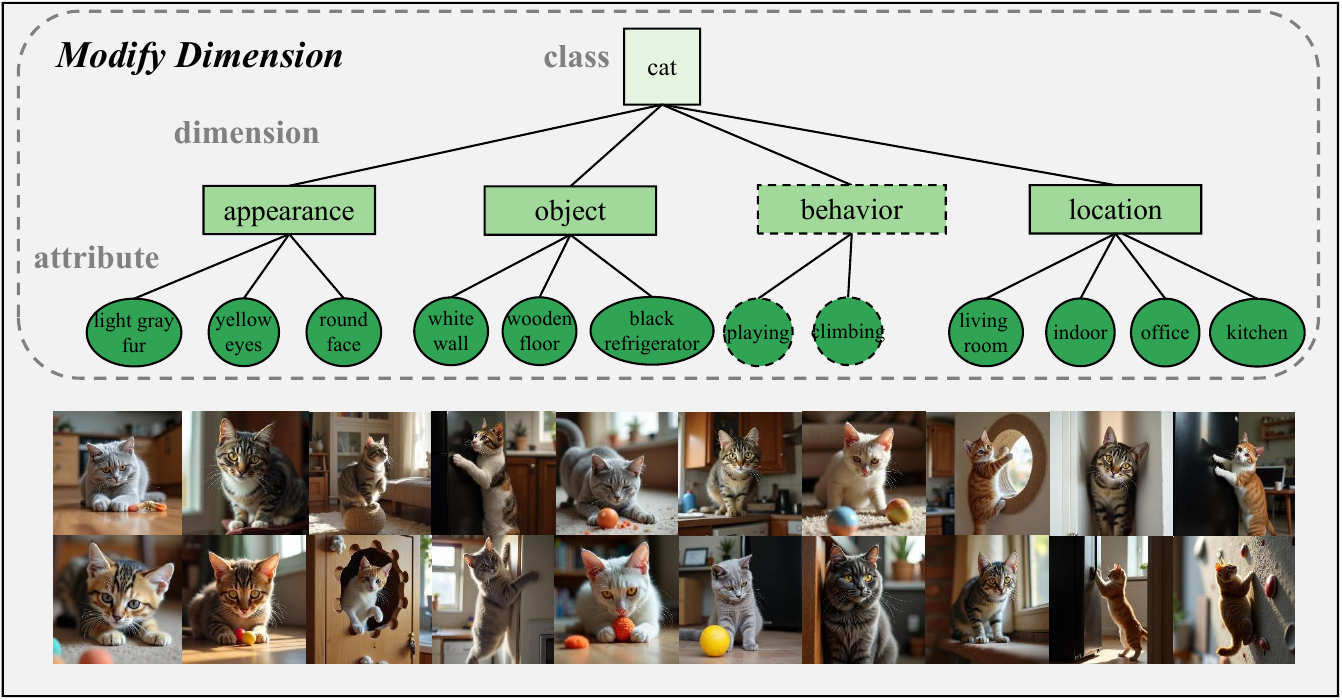}
    \caption{By changing the ``action" dimension in the original concept tree to the ``behavior" dimension, cats are no longer just squatting or lying down, bringing about a variety of posture changes.}
    \label{fig:cat_modify}
\end{figure*}

\begin{figure*}[ht]
    \centering
    \includegraphics[width=\textwidth]{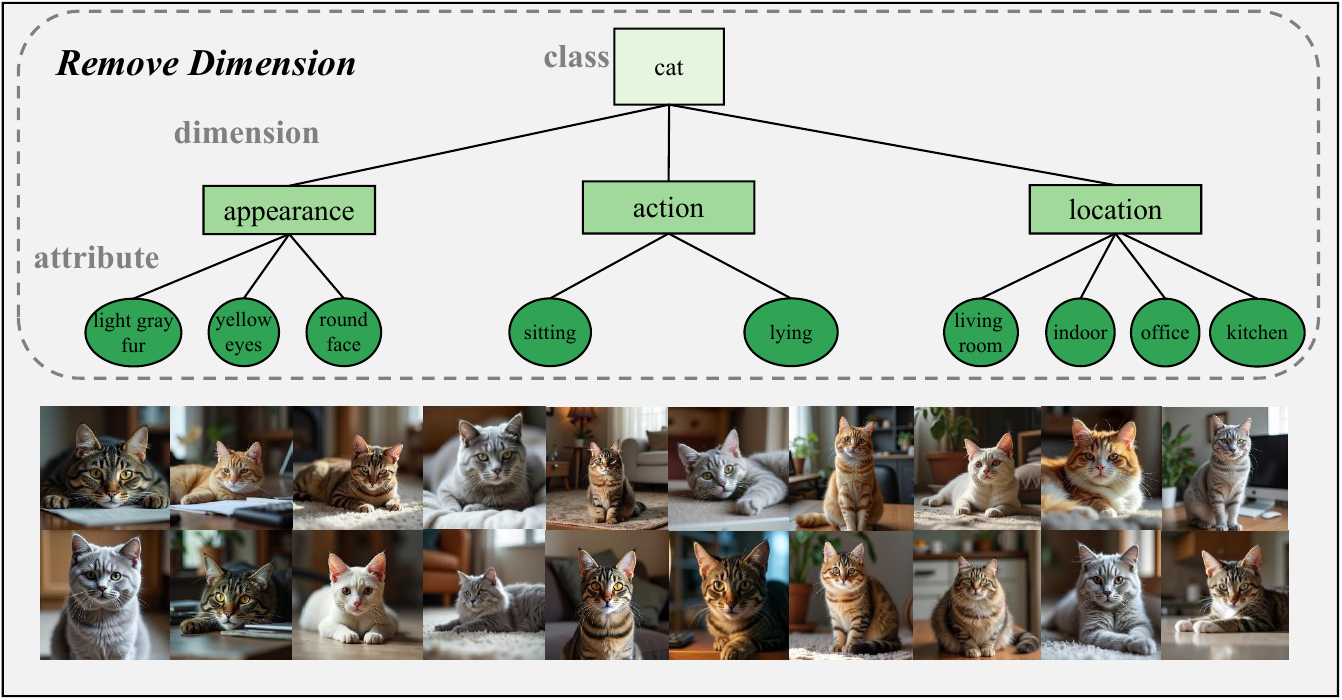}
    \caption{Removing the ``object" dimension from the original concept tree makes the objects around the cat become monotonous. The diversity of images has significantly decreased.}
    \label{fig:cat_remove}
\end{figure*}

\begin{figure*}[ht]
    \centering
    \includegraphics[width=\textwidth]{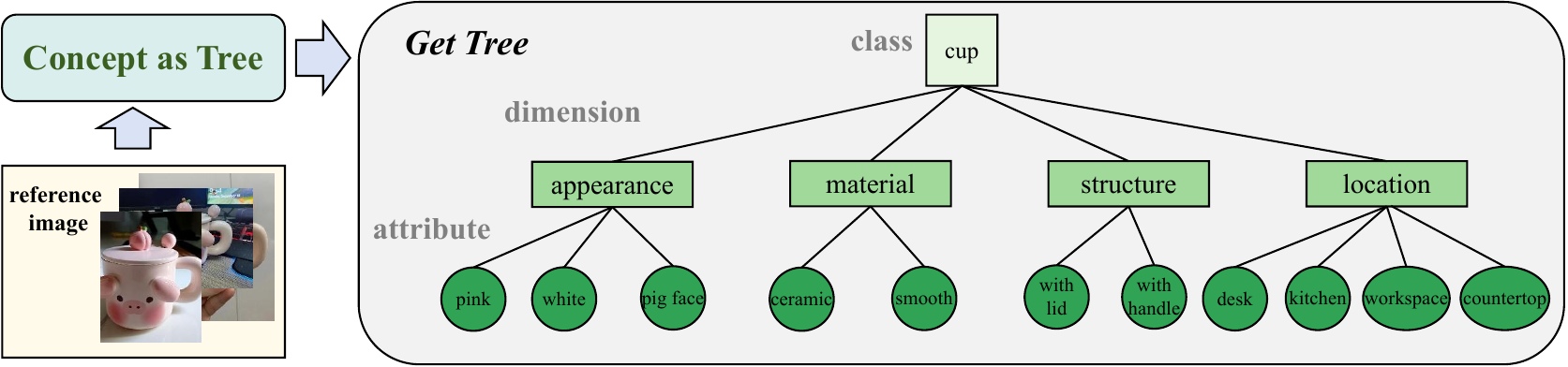}
    \caption{Using the CaT framework, we generate a concept tree for the ``cup" concept. The CaT framework can accurately extract the visual information associated with this concept, such as material and structure.}
    \label{fig:obj_get}
\end{figure*}

\begin{figure*}[ht]
    \centering
    \includegraphics[width=\textwidth]{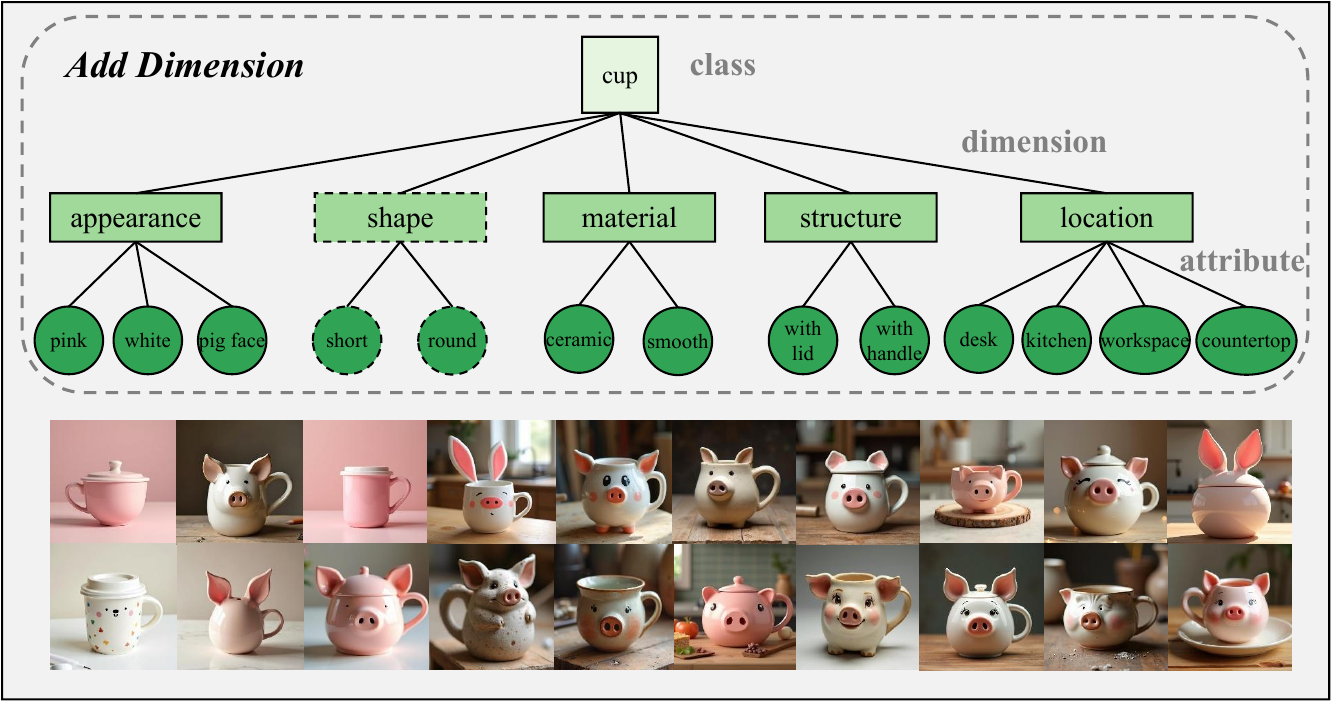}
    \caption{Adding a ``shape" dimension to the original concept tree, the height and weight of the cups are no longer consistent, bringing about an increase in diversity.}
    \label{fig:obj_add}
\end{figure*}

\begin{figure*}[ht]
    \centering
    \includegraphics[width=\textwidth]{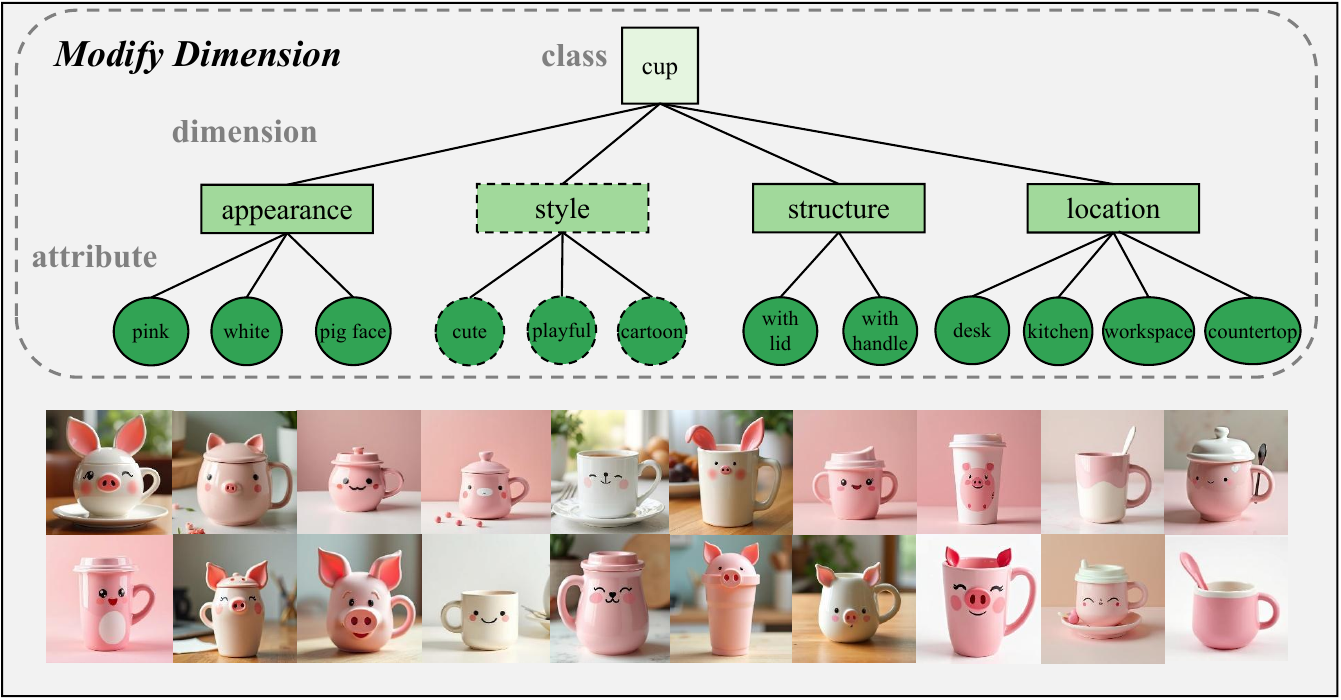}
    \caption{Changing the ``material" dimension in the original concept tree to the ``style" dimension makes the expression on the cup more vivid and adorable.}
    \label{fig:obj_modify}
\end{figure*}

\begin{figure*}[ht]
    \centering
    \includegraphics[width=\textwidth]{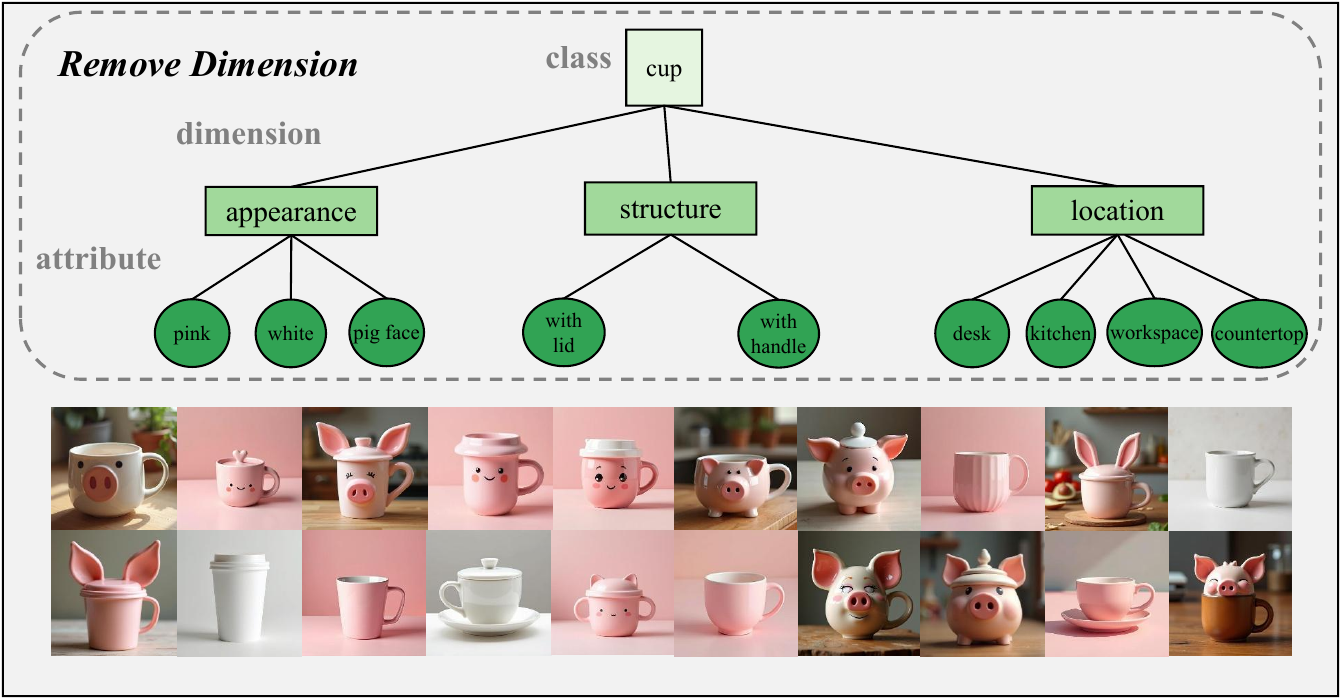}
    \caption{After removing the ``material" dimension from the original concept tree, most of the generated images only have one material, smoothness. The diversity of images has significantly decreased.}
    \label{fig:obj_remove}
\end{figure*}

\begin{figure*}[t]
    \centering
    \includegraphics[width=\textwidth]{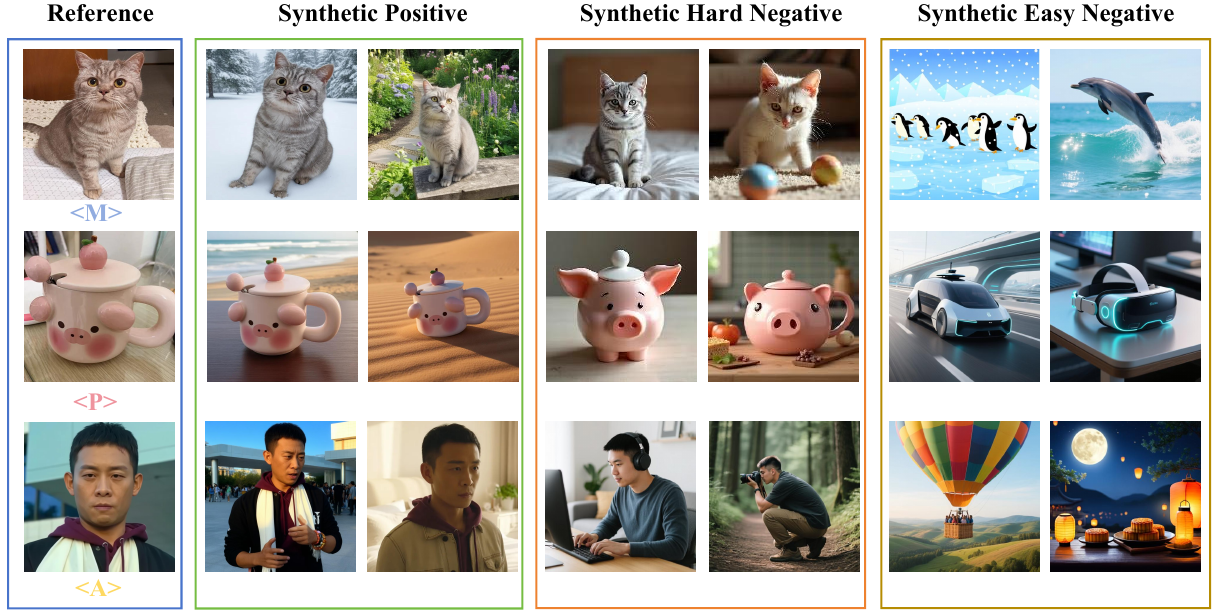}
    \caption{Example positive, easy negative, and hard negative samples from multiple concepts.}
    \label{fig:supple_training_data}
\end{figure*}

We provide detailed visualizations of Concept Trees, covering three categories: humans in ~\cref{fig:man_get,fig:man_add,fig:man_modify,fig:man_remove,fig:woman_get,fig:woman_add,fig:woman_modify,fig:woman_remove}, pets in~\cref{fig:cat_get,fig:cat_add,fig:cat_modify,fig:cat_remove}, and objects in ~\cref{fig:obj_get,fig:obj_add,fig:obj_modify,fig:obj_remove}.
For each concept, we first obtain its concept tree using the CaT. Subsequently, we modify the original tree through three types of tree editing operations: add, modify, and remove. After each operation, a new concept tree is generated, which is then used to synthesize diverse personalized images. 
In addition, we provide detailed positive, easy negative, and hard negative samples for some concepts, as shown in Fig.~\ref{fig:supple_training_data}.

\section{CaT Prompt}
\label{sec:cat_prompt}
We provide all the prompts used in the CaT framework. 
Tab.~\ref{tab:get_tree} includes the three steps for concept tree generation: obtaining the reference image description, performing batch summarization to generate the initial concept tree, and then refining it through self-refinement to obtain the final, well-developed concept tree. 
Tab.~\ref{tab:get_easy_tree} presents the initialization prompts for the simple negative sample concept tree. 
Finally, Tab.~\ref{tab:tree_editing} provides the three editing operations for the tree—addition, removal, and modification—as well as the prompts used for synthesizing text for image generation.

\section{Human Evaluation}
\label{sec:human_eval}
In addition to the various experiments in the main text that demonstrate the diversity and effectiveness of our synthesized positive and hard negative samples, we conducted a site-by-site evaluation experiment comparing synthetic data with original data. 
The evaluators were college student volunteers recruited by us. Our evaluation objective was to demonstrate that the synthesized positive and hard negative samples have high quality to the original concept images.

The original data were sourced from MyVLM, Yo'llaVA, and MC-LLaVA. For each concept, we selected 1 to 3 positive samples as reference images. We then conducted site-by-site evaluation experiments for both positive and negative samples. 
For the original positive samples (excluding those selected as reference images) and synthesized positive samples, we randomly paired one from each group and compared them sequentially with all reference images of the corresponding concept. 
Evaluators were required to choose the image with higher similarity to the reference image and assign it one point. The winner of each comparison was determined based on the cumulative scores across all reference images. 
The same evaluation method was applied to the original retrieved complex negative samples and synthesized hard negative samples. Note that if the number of reference images was even, there was a possibility of a tie.

\begin{figure}[!ht]
    \centering
    \includegraphics[width=0.8\textwidth]{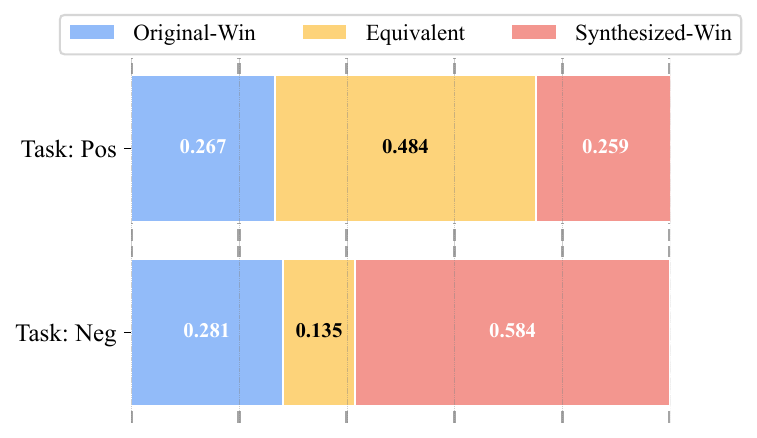}
    \caption{Human evaluation of similarity between original and synthesized images compared to reference images: The synthesized positive samples were found to be on par with the original positive samples, while the synthesized hard negative samples demonstrated higher similarity compared to the originally retrieved hard negative samples.}
    \label{fig:human_eval}
\end{figure}

The results of the two sets of experiments are presented in Fig.~\ref{fig:human_eval}. We find that the synthesized positive samples perform comparably to the original positive samples. Meanwhile, the synthesized hard negative samples achieve significantly higher scores than the retrieved hard negative samples. 
This result clearly demonstrates that our synthesized data can effectively support the personalization training.

\section{Limitation and Future Work}
\label{sec:limitation}
Although our method demonstrates effectiveness in generating concept-centric data, several limitations warrant consideration, along with potential directions for future work.

The first limitation stems from our framework's current inability to handle scenarios involving multiple concepts, primarily due to its dependence on FLUX~\cite{flux2024} for image generation. Nevertheless, we believe the inherent scalability of our tree structure paradigm offers promising opportunities for extension through forest-based architectures. Additionally, Future work could explore the integration of enhanced diffusion models with stronger guidance mechanisms, such as layout or box conditions, which have shown potential in capturing multiple concepts information.

A second limitation arises from potential biases inherited from the LAION-5B~\cite{schuhmann2022laion5bopenlargescaledataset} dataset used for training FLUX. However, this issue is actively addressed in ongoing research. Recent studies~\cite{seshadri2023bias, friedrich2023fair, su2023manifold} have made significant progress in developing techniques to enhance fairness and reduce biases in generative models, providing directions for future improvements in our framework.

Finally, the generated concept-centric images may bring data privacy and ethical concerns. This challenge has garnered increasing attention in the current research community, with recent works~\cite{he2024diff, you2024generation,liu2023diffprotect} making notable advances in privacy-preserving generation techniques. We anticipate that continued research in GenAI safety will yield effective solutions to mitigate these concerns, enabling more responsible use of synthetic data generation methods.

\begin{table*}[ht]\centering
\begin{minipage}{0.95\textwidth}
\centering
\begin{tcolorbox}[colback=verylightgray,boxrule=1.2pt]
\small
\textcolor{blue!90}{\textit{\textbf{Image Description.}}}\newline
Please identify the primary object class name in the image and describe the image in detail with the class name. \newline

\textcolor{blue!90}{\textit{\textbf{Batch Summarization.}}}\newline
\textbf{Task}: I want to classify and organize captions for some images \newline
\textbf{Requirement}: I will provide a batch of captions and the main object they revolve around. Please describe the attributes related to the subject from multiple dimensions based on the subtitles. 
The final result is output as \{tree\_example\}\newline
\textbf{Input}: main object: \{class\_name\}; captions: \{captions\}\newline
\textbf{Output}:\newline

\textcolor{blue!90}{\textit{\textbf{Self-Refinement.}}}\newline
\textbf{Task}: I'm trying to classify the following captions into a classification criteria. However, it seems the current criteria fails to capture certain details that would help differentiate this caption.\newline
\textbf{Requirement}: We are unable to classify these captions using the provided criteria due to one of the following reasons.\newline
1. LLM Hallucination: If you believe the current criteria is reasonable, and the sample can be classified under one of them, the current failure may be due to LLM hallucination. If the majority of classifications are correct and only a small portion of the results appears highly unreasonable, it is likely due to hallucination. In this case, please do nothing. Answer format: \{\{``hallucination": []\}\}\newline
2. Attribute Redundancy: If there are redundant attributes in the criteria, please identify and replace them with a single, unified keyword that represents all the redundant attributes. Answer format: \{\{``redundant": [``unified keyword"]\}\}
(Only include a single keyword that replaces all the redundant or duplicated attributes).\newline
3. Missing Attributes: If some important attributes are missing and need to be added to the criteria to accurately classify the caption, please suggest one attribute to add.
Answer format: \{\{``missing": [``keyword"]\}\}\newline
After obtaining the content in the curly braces (for example, {{``hallucination": []}}, {{``redundant": [...]}}, {{``missing": [...]}}), please do the following processing according to the situation.\newline
1. For hallucination, it means that the previous LLM judgment is wrong, and there is no need to modify the current tree;\newline
2. For redundancy, please merge them and put the final attribute into a dimension you think is appropriate (if necessary, you can delete the redundant dimension);\newline
3. For missing, please find the key information in the content of captions and extract the missing attributes, and add them to the appropriate dimension (if necessary, you can create a new dimension)\newline
\textbf{Input}: caption: \{captions\}; 
current concept tree: \{concept\_tree\} \newline
\textbf{Output}:
\end{tcolorbox}
\caption{The three steps of concept tree synthesis.}
    \label{tab:get_tree}
\end{minipage}
\end{table*}

\begin{table*}[ht]\centering
\begin{minipage}{0.95\textwidth}
\centering
\begin{tcolorbox}[colback=verylightgray,boxrule=1.2pt]
\small
\textcolor{blue!90}{\textit{\textbf{Easy Negative Sample Concept Tree Generation.}}}\newline
\textbf{Task}: I want to modify the visual definition of a class to synthesize a new visual definition of the class\newline
\textbf{Requirement}: I will input the visual definition tree of a class. Please modify its class name to obtain a new class, and generate some visual dimensions and corresponding attributes that match the new class according to the tree format. Finally, output the visual definition tree in the original tree format to return the new visual definition tree\newline
\textbf{Input}: concept tree: \{concept\_tree\}\newline
\textbf{Output}:
\end{tcolorbox}
\caption{Modify the root node of the original concept tree to obtain an easy negative sample concept tree.}
    \label{tab:get_easy_tree}
\end{minipage}
\end{table*}

\begin{table*}[ht]\centering
\begin{minipage}{0.95\textwidth}
\centering
\begin{tcolorbox}[colback=verylightgray,boxrule=1.2pt]
\small
\textcolor{blue!90}{\textit{\textbf{Tree Operation---Add.}}}\newline
\textbf{Task}: I want to modify the visual definition tree of a class\newline
\textbf{Requirement}: I will input a visual definition tree for a class, please randomly add \{num\} of its dimensions to obtain a new visual definition tree, note that it conforms to the natural definition of this class. Only output the new visual definition tree.\newline
\textbf{Input}: concept tree: \{concept\_tree\}\newline
\textbf{Output}:\newline

\textcolor{blue!90}{\textit{\textbf{Tree Operation---Remove.}}}\newline
\textbf{Task}: I want to modify the visual definition tree of a class\newline
\textbf{Requirement}: I will input a visual definition tree for a class, please randomly delete \{num\} of its dimensions to obtain a new visual definition tree, note that it conforms to the natural definition of this class. Only output the new visual definition tree.\newline
\textbf{Input}: concept tree: \{concept\_tree\}\newline
\textbf{Output}:\newline

\textcolor{blue!90}{\textit{\textbf{Tree Operation---Modify.}}}\newline
\textbf{Task}: I want to modify the visual definition tree of a class\newline
\textbf{Requirement}: I will input a visual definition tree for a class, please randomly modify \{num\} of its dimensions to obtain a new visual definition tree, it means deleting some visual dimensions and adding new visual dimensions, not just modifying attributes. Only output the new visual definition tree.\newline
\textbf{Input}: concept tree: \{concept\_tree\}\newline
\textbf{Output}:\newline

\textcolor{blue!90}{\textit{\textbf{Image Prompt Generation.}}}\newline
\textbf{Task}: I want to generate different prompts based on the visual definition of a category to prompt the diffusion model to synthesize images.\newline
\textbf{Requirement}: I will provide a visual definition of a category, including the category name and its multiple visual dimensions and attributes. You can combine the attributes of these dimensions to obtain prompts that match the real scene, such as ``a photo of attribute of dimension1, attribute of dimension2, ..., classname". Please generate at least 100 prompts, ensuring that the similarity of each prompt is 0, and only output prompts line by line.\newline
\textbf{Input}: class category: \{category\}; concept tree: \{concept\_tree\}\newline
\textbf{Output}:
\end{tcolorbox}
\caption{Three editing operations of a concept tree and using a concept tree to synthesize diverse image prompts.}
    \label{tab:tree_editing}
\end{minipage}
\end{table*}

\clearpage

\clearpage

%% file: main.bib
@String(CVPR= {IEEE Conf. Comput. Vis. Pattern Recog.})

@String(CVPR  = {CVPR})

@article{bai2023qwenvl,
  title={Qwen-vl: A frontier large vision-language model with versatile abilities},
  author={Bai, Jinze and Bai, Shuai and Yang, Shusheng and Wang, Shijie and Tan, Sinan and Wang, Peng and Lin, Junyang and Zhou, Chang and Zhou, Jingren},
  journal={arXiv preprint arXiv:2308.12966},
  year={2023}
}

@article{lin2023sphinx,
  title={Sphinx: The joint mixing of weights, tasks, and visual embeddings for multi-modal large language models},
  author={Lin, Ziyi and Liu, Chris and Zhang, Renrui and Gao, Peng and Qiu, Longtian and Xiao, Han and Qiu, Han and Lin, Chen and Shao, Wenqi and Chen, Keqin and others},
  journal={arXiv preprint arXiv:2311.07575},
  year={2023}
}

@inproceedings{li2023blip,
  title={Blip-2: Bootstrapping language-image pre-training with frozen image encoders and large language models},
  author={Li, Junnan and Li, Dongxu and Savarese, Silvio and Hoi, Steven},
  booktitle={International conference on machine learning},
  pages={19730--19742},
  year={2023},
  organization={PMLR}
}

@article{achiam2023gpt,
  title={Gpt-4 technical report},
  author={Achiam, Josh and Adler, Steven and Agarwal, Sandhini and Ahmad, Lama and Akkaya, Ilge and Aleman, Florencia Leoni and Almeida, Diogo and Altenschmidt, Janko and Altman, Sam and Anadkat, Shyamal and others},
  journal={arXiv preprint arXiv:2303.08774},
  year={2023}
}

@article{lin2024draw,
  title={Draw-and-Understand: Leveraging Visual Prompts to Enable MLLMs to Comprehend What You Want},
  author={Lin, Weifeng and Wei, Xinyu and An, Ruichuan and Gao, Peng and Zou, Bocheng and Luo, Yulin and Huang, Siyuan and Zhang, Shanghang and Li, Hongsheng},
  journal={arXiv preprint arXiv:2403.20271},
  year={2024}
}

@article{luo2024llm,
  title={LLM as Dataset Analyst: Subpopulation Structure Discovery with Large Language Model},
  author={Luo, Yulin and An, Ruichuan and Zou, Bocheng and Tang, Yiming and Liu, Jiaming and Zhang, Shanghang},
  journal={arXiv preprint arXiv:2405.02363},
  year={2024}
}

@article{an2025unictokens,
  title={UniCTokens: Boosting Personalized Understanding and Generation via Unified Concept Tokens},
  author={An, Ruichuan and Yang, Sihan and Zhang, Renrui and Shen, Zijun and Lu, Ming and Dai, Gaole and Liang, Hao and Guo, Ziyu and Yan, Shilin and Luo, Yulin and others},
  journal={arXiv preprint arXiv:2505.14671},
  year={2025}
}

@inproceedings{hao2025rap,
  title={RAP: Retrieval-Augmented Personalization for Multimodal Large Language Models},
  author={Hao, Haoran and Han, Jiaming and Li, Changsheng and Li, Yu-Feng and Yue, Xiangyu},
  booktitle={Proceedings of the Computer Vision and Pattern Recognition Conference},
  pages={14538--14548},
  year={2025}
}

@article{cao2025move,
  title={MoVE-KD: Knowledge Distillation for VLMs with Mixture of Visual Encoders},
  author={Cao, Jiajun and Zhang, Yuan and Huang, Tao and Lu, Ming and Zhang, Qizhe and An, Ruichuan and Ma, Ningning and Zhang, Shanghang},
  journal={arXiv preprint arXiv:2501.01709},
  year={2025}
}

@inproceedings{alaluf2025myvlm,
  title={Myvlm: Personalizing vlms for user-specific queries},
  author={Alaluf, Yuval and Richardson, Elad and Tulyakov, Sergey and Aberman, Kfir and Cohen-Or, Daniel},
  booktitle={European Conference on Computer Vision},
  pages={73--91},
  year={2025},
  organization={Springer}
}

@article{nguyenyo,
  title={Yo'LLaVA: Your Personalized Language and Vision Assistant},
  author={Nguyen, Thao and Liu, Haotian and Li, Yuheng and Cai, Mu and Ojha, Utkarsh and Lee, Yong Jae},
  journal={arXiv preprint arXiv:2406.09400},
  year={2024}
}

@misc{an2024mcllavamulticonceptpersonalizedvisionlanguage,
      title={MC-LLaVA: Multi-Concept Personalized Vision-Language Model}, 
      author={Ruichuan An and Sihan Yang and Ming Lu and Kai Zeng and Yulin Luo and Ying Chen and Jiajun Cao and Hao Liang and Qi She and Shanghang Zhang and Wentao Zhang},
      year={2024},
      eprint={2411.11706},
      archivePrefix={arXiv},
      primaryClass={cs.CV},
      url={https://arxiv.org/abs/2411.11706}
}

@misc{hao2024rememberretrievegenerateunderstanding,
        title={Remember, Retrieve and Generate: Understanding Infinite Visual Concepts as Your Personalized Assistant}, 
        author={Haoran Hao and Jiaming Han and Changsheng Li and Yu-Feng Li and Xiangyu Yue},
        year={2024},
        eprint={2410.13360},
        archivePrefix={arXiv},
        primaryClass={cs.CV},
        url={https://arxiv.org/abs/2410.13360}
}

@article{song2021bob,
  title={BoB: BERT over BERT for training persona-based dialogue models from limited personalized data},
  author={Song, Haoyu and Wang, Yan and Zhang, Kaiyan and Zhang, Wei-Nan and Liu, Ting},
  journal={arXiv preprint arXiv:2106.06169},
  year={2021}
}

@inproceedings{welch2022leveraging,
  title={Leveraging similar users for personalized language modeling with limited data},
  author={Welch, Charles and Gu, Chenxi and Kummerfeld, Jonathan K and Perez-Rosas, Veronica and Mihalcea, Rada},
  booktitle={Proceedings of the 60th Annual Meeting of the Association for Computational Linguistics (Volume 1: Long Papers)},
  pages={1742--1752},
  year={2022}
}

@article{chawla2013bringing,
  title={Bringing big data to personalized healthcare: a patient-centered framework},
  author={Chawla, Nitesh V and Davis, Darcy A},
  journal={Journal of general internal medicine},
  volume={28},
  pages={660--665},
  year={2013},
  publisher={Springer}
}

@article{meghwani2024enhancing,
  title={Enhancing Retrieval Performance: An Ensemble Approach For Hard Negative Mining},
  author={Meghwani, Hansa},
  journal={arXiv preprint arXiv:2411.02404},
  year={2024}
}

@article{liu2023visual,
  title={Visual instruction tuning},
  author={Liu, Haotian and Li, Chunyuan and Wu, Qingyang and Lee, Yong Jae},
  journal={Advances in neural information processing systems},
  volume={36},
  pages={34892--34916},
  year={2023}
}

@article{li2024llava,
  title={Llava-onevision: Easy visual task transfer},
  author={Li, Bo and Zhang, Yuanhan and Guo, Dong and Zhang, Renrui and Li, Feng and Zhang, Hao and Zhang, Kaichen and Zhang, Peiyuan and Li, Yanwei and Liu, Ziwei and others},
  journal={arXiv preprint arXiv:2408.03326},
  year={2024}
}

@article{li2024llava_next,
  title={Llava-next-interleave: Tackling multi-image, video, and 3d in large multimodal models},
  author={Li, Feng and Zhang, Renrui and Zhang, Hao and Zhang, Yuanhan and Li, Bo and Li, Wei and Ma, Zejun and Li, Chunyuan},
  journal={arXiv preprint arXiv:2407.07895},
  year={2024}
}

@article{dorka2024training,
  title={Training a vision language model as smartphone assistant},
  author={Dorka, Nicolai and Marecki, Janusz and Anwar, Ammar},
  journal={arXiv preprint arXiv:2404.08755},
  year={2024}
}

@article{zhang2024vision,
  title={Vision Search Assistant: Empower Vision-Language Models as Multimodal Search Engines},
  author={Zhang, Zhixin and Zhang, Yiyuan and Ding, Xiaohan and Yue, Xiangyu},
  journal={arXiv preprint arXiv:2410.21220},
  year={2024}
}

@article{an2024mc,
  title={MC-LLaVA: Multi-Concept Personalized Vision-Language Model},
  author={An, Ruichuan and Yang, Sihan and Lu, Ming and Zeng, Kai and Luo, Yulin and Chen, Ying and Cao, Jiajun and Liang, Hao and She, Qi and Zhang, Shanghang and others},
  journal={arXiv preprint arXiv:2411.11706},
  year={2024}
}

@article{jordon2022synthetic,
  title={Synthetic Data--what, why and how?},
  author={Jordon, James and Szpruch, Lukasz and Houssiau, Florimond and Bottarelli, Mirko and Cherubin, Giovanni and Maple, Carsten and Cohen, Samuel N and Weller, Adrian},
  journal={arXiv preprint arXiv:2205.03257},
  year={2022}
}

@article{long2024llms,
  title={On llms-driven synthetic data generation, curation, and evaluation: A survey},
  author={Long, Lin and Wang, Rui and Xiao, Ruixuan and Zhao, Junbo and Ding, Xiao and Chen, Gang and Wang, Haobo},
  journal={arXiv preprint arXiv:2406.15126},
  year={2024}
}

@article{guo2024generative,
  title={Generative ai for synthetic data generation: Methods, challenges and the future},
  author={Guo, Xu and Chen, Yiqiang},
  journal={arXiv preprint arXiv:2403.04190},
  year={2024}
}

@inproceedings{shipard2023diversity,
  title={Diversity is definitely needed: Improving model-agnostic zero-shot classification via stable diffusion},
  author={Shipard, Jordan and Wiliem, Arnold and Thanh, Kien Nguyen and Xiang, Wei and Fookes, Clinton},
  booktitle={Proceedings of the IEEE/CVF Conference on Computer Vision and Pattern Recognition},
  pages={769--778},
  year={2023}
}

@article{yang2023freemask,
  title={Freemask: Synthetic images with dense annotations make stronger segmentation models},
  author={Yang, Lihe and Xu, Xiaogang and Kang, Bingyi and Shi, Yinghuan and Zhao, Hengshuang},
  journal={Advances in Neural Information Processing Systems},
  volume={36},
  pages={18659--18675},
  year={2023}
}

@inproceedings{feng2024instagen,
  title={Instagen: Enhancing object detection by training on synthetic dataset},
  author={Feng, Chengjian and Zhong, Yujie and Jie, Zequn and Xie, Weidi and Ma, Lin},
  booktitle={Proceedings of the IEEE/CVF Conference on Computer Vision and Pattern Recognition},
  pages={14121--14130},
  year={2024}
}

@article{liu2024synthvlm,
  title={Synthvlm: High-efficiency and high-quality synthetic data for vision language models},
  author={Liu, Zheng and Liang, Hao and Huang, Xijie and Xiong, Wentao and Yu, Qinhan and Sun, Linzhuang and Chen, Chong and He, Conghui and Cui, Bin and Zhang, Wentao},
  journal={arXiv preprint arXiv:2407.20756},
  year={2024}
}

@article{liang2024synth,
  title={Synth-empathy: Towards high-quality synthetic empathy data},
  author={Liang, Hao and Sun, Linzhuang and Wei, Jingxuan and Huang, Xijie and Sun, Linkun and Yu, Bihui and He, Conghui and Zhang, Wentao},
  journal={arXiv preprint arXiv:2407.21669},
  year={2024}
}

@article{fan2023improving,
  title={Improving clip training with language rewrites},
  author={Fan, Lijie and Krishnan, Dilip and Isola, Phillip and Katabi, Dina and Tian, Yonglong},
  journal={Advances in Neural Information Processing Systems},
  volume={36},
  pages={35544--35575},
  year={2023}
}

@inproceedings{peng2024synthesize,
  title={Synthesize diagnose and optimize: Towards fine-grained vision-language understanding},
  author={Peng, Wujian and Xie, Sicheng and You, Zuyao and Lan, Shiyi and Wu, Zuxuan},
  booktitle={Proceedings of the IEEE/CVF Conference on Computer Vision and Pattern Recognition},
  pages={13279--13288},
  year={2024}
}

@article{li2024synthetic,
  title={Synthetic data (almost) from scratch: Generalized instruction tuning for language models},
  author={Li, Haoran and Dong, Qingxiu and Tang, Zhengyang and Wang, Chaojun and Zhang, Xingxing and Huang, Haoyang and Huang, Shaohan and Huang, Xiaolong and Huang, Zeqiang and Zhang, Dongdong and others},
  journal={arXiv preprint arXiv:2402.13064},
  year={2024}
}

@article{dunlap2023diversify,
  title={Diversify your vision datasets with automatic diffusion-based augmentation},
  author={Dunlap, Lisa and Umino, Alyssa and Zhang, Han and Yang, Jiezhi and Gonzalez, Joseph E and Darrell, Trevor},
  journal={Advances in neural information processing systems},
  volume={36},
  pages={79024--79034},
  year={2023}
}

@article{zeng2026steervte,
  title={SteerVTE: Seamless Video Text Editing with Style and Glyph Control},
  author={Zeng, Kai and Li, Moran and Wang, Zhengwei and Yu, Yingchen and Lin, Yiheng and An, Ruichuan and Lu, Ming and She, Qi and Zhang, Wentao},
  journal={arXiv preprint arXiv:2606.23254},
  year={2026}
}

@article{wang2024template,
  title={Template Matters: Understanding the Role of Instruction Templates in Multimodal Language Model Evaluation and Training},
  author={Wang, Shijian and Song, Linxin and Zhang, Jieyu and Shimizu, Ryotaro and Luo, Ao and Yao, Li and Chen, Cunjian and McAuley, Julian and Wu, Hanqian},
  journal={arXiv preprint arXiv:2412.08307},
  year={2024}
}

@misc{huang2024modifyingdataaddressgraph,
      title={Can Modifying Data Address Graph Domain Adaptation?}, 
      author={Renhong Huang and Jiarong Xu and Xin Jiang and Ruichuan An and Yang Yang},
      year={2024},
      eprint={2407.19311},
      archivePrefix={arXiv},
      primaryClass={cs.LG},
      url={https://arxiv.org/abs/2407.19311}, 
}

@inproceedings{
wang2024attributed,
title={Attributed Synthetic Data Generation for Zero-shot Image Classification},
author={Shijian Wang and Linxin Song and Ryotaro Shimizu and Masayuki Goto and Hanqian wu},
booktitle={Synthetic Data for Computer Vision Workshop @ CVPR 2024},
year={2024},
url={https://openreview.net/forum?id=k4Xnh0EPus}
}

@article{dong2025scalable,
  title={Scalable vision language model training via high quality data curation},
  author={Dong, Hongyuan and Kang, Zijian and Yin, Weijie and Liang, Xiao and Feng, Chao and Ran, Jiao},
  journal={arXiv preprint arXiv:2501.05952},
  year={2025}
}

@inproceedings{goyal2024scaling,
  title={Scaling Laws for Data Filtering--Data Curation cannot be Compute Agnostic},
  author={Goyal, Sachin and Maini, Pratyush and Lipton, Zachary C and Raghunathan, Aditi and Kolter, J Zico},
  booktitle={Proceedings of the IEEE/CVF Conference on Computer Vision and Pattern Recognition},
  pages={22702--22711},
  year={2024}
}

@article{ridzuan2024review,
  title={A Review on Data Quality Dimensions for Big Data},
  author={Ridzuan, Fakhitah and Zainon, Wan Mohd Nazmee Wan},
  journal={Procedia Computer Science},
  volume={234},
  pages={341--348},
  year={2024},
  publisher={Elsevier}
}

@article{chen2024your,
  title={Your vision-language model itself is a strong filter: Towards high-quality instruction tuning with data selection},
  author={Chen, Ruibo and Wu, Yihan and Chen, Lichang and Liu, Guodong and He, Qi and Xiong, Tianyi and Liu, Chenxi and Guo, Junfeng and Huang, Heng},
  journal={arXiv preprint arXiv:2402.12501},
  year={2024}
}

@article{yi2025bridge,
  title={Bridge the Modality and Capability Gaps in Vision-Language Model Selection},
  author={Yi, Chao and He, Yuhang and Zhan, De-Chuan and Ye, Han-Jia},
  journal={Advances in Neural Information Processing Systems},
  volume={37},
  pages={34429--34452},
  year={2025}
}

@article{awotunde2021intrusion,
  title={Intrusion detection in industrial internet of things network-based on deep learning model with rule-based feature selection},
  author={Awotunde, Joseph Bamidele and Chakraborty, Chinmay and Adeniyi, Abidemi Emmanuel},
  journal={Wireless communications and mobile computing},
  volume={2021},
  number={1},
  pages={7154587},
  year={2021},
  publisher={Wiley Online Library}
}

@article{fabris2025efficient,
  title={Efficient sensors selection for traffic flow monitoring: An overview of model-based techniques leveraging network observability},
  author={Fabris, Marco and Ceccato, Riccardo and Zanella, Andrea},
  journal={Sensors},
  volume={25},
  number={5},
  pages={1416},
  year={2025},
  publisher={MDPI}
}

@article{zeng2025rethinking,
  title={Rethinking Driving World Model as Synthetic Data Generator for Perception Tasks},
  author={Zeng, Kai and Wu, Zhanqian and Xiong, Kaixin and Wei, Xiaobao and Guo, Xiangyu and Zhu, Zhenxin and Ho, Kalok and Zhou, Lijun and Zeng, Bohan and Lu, Ming and others},
  journal={arXiv preprint arXiv:2510.19195},
  year={2025}
}

@inproceedings{hong2024s,
  title={Who's in and who's out? A case study of multimodal CLIP-filtering in DataComp},
  author={Hong, Rachel and Agnew, William and Kohno, Tadayoshi and Morgenstern, Jamie},
  booktitle={Proceedings of the 4th ACM Conference on Equity and Access in Algorithms, Mechanisms, and Optimization},
  pages={1--17},
  year={2024}
}

@article{wang2024finetuned,
  title={Finetuned multimodal language models are high-quality image-text data filters},
  author={Wang, Weizhi and Mrini, Khalil and Yang, Linjie and Kumar, Sateesh and Tian, Yu and Yan, Xifeng and Wang, Heng},
  journal={arXiv preprint arXiv:2403.02677},
  year={2024}
}

@inproceedings{radford2021learning,
  title={Learning transferable visual models from natural language supervision},
  author={Radford, Alec and Kim, Jong Wook and Hallacy, Chris and Ramesh, Aditya and Goh, Gabriel and Agarwal, Sandhini and Sastry, Girish and Askell, Amanda and Mishkin, Pamela and Clark, Jack and others},
  booktitle={International conference on machine learning},
  pages={8748--8763},
  year={2021},
  organization={PmLR}
}

@inproceedings{ruiz2023dreambooth,
  title={Dreambooth: Fine tuning text-to-image diffusion models for subject-driven generation},
  author={Ruiz, Nataniel and Li, Yuanzhen and Jampani, Varun and Pritch, Yael and Rubinstein, Michael and Aberman, Kfir},
  booktitle={Proceedings of the IEEE/CVF conference on computer vision and pattern recognition},
  pages={22500--22510},
  year={2023}
}

@misc{flux2024,
    author={Black Forest Labs},
    title={FLUX},
    year={2024},
    howpublished={\url{https://github.com/black-forest-labs/flux}},
}

@misc{hong2025dialoguelanguagemodellargescale,
      title={Dialogue Language Model with Large-Scale Persona Data Engineering}, 
      author={Mengze Hong and Chen Jason Zhang and Chaotao Chen and Rongzhong Lian and Di Jiang},
      year={2025},
      eprint={2412.09034},
      archivePrefix={arXiv},
      primaryClass={cs.CL},
      url={https://arxiv.org/abs/2412.09034}, 
}

@misc{schuhmann2022laion5bopenlargescaledataset,
      title={LAION-5B: An open large-scale dataset for training next generation image-text models}, 
      author={Christoph Schuhmann and Romain Beaumont and Richard Vencu and Cade Gordon and Ross Wightman and Mehdi Cherti and Theo Coombes and Aarush Katta and Clayton Mullis and Mitchell Wortsman and Patrick Schramowski and Srivatsa Kundurthy and Katherine Crowson and Ludwig Schmidt and Robert Kaczmarczyk and Jenia Jitsev},
      year={2022},
      eprint={2210.08402},
      archivePrefix={arXiv},
      primaryClass={cs.CV},
      url={https://arxiv.org/abs/2210.08402}, 
}

@article{he2024diff,
  title={Diff-privacy: Diffusion-based face privacy protection},
  author={He, Xiao and Zhu, Mingrui and Chen, Dongxin and Wang, Nannan and Gao, Xinbo},
  journal={IEEE Transactions on Circuits and Systems for Video Technology},
  year={2024},
  publisher={IEEE}
}

@article{you2024generation,
  title={Generation of Face Privacy-Protected Images Based on the Diffusion Model},
  author={You, Xingyi and Zhao, Xiaohu and Wang, Yue and Sun, Weiqing},
  journal={Entropy},
  volume={26},
  number={6},
  pages={479},
  year={2024},
  publisher={MDPI}
}

@article{liu2023diffprotect,
  title={Diffprotect: Generate adversarial examples with diffusion models for facial privacy protection},
  author={Liu, Jiang and Lau, Chun Pong and Chellappa, Rama},
  journal={arXiv preprint arXiv:2305.13625},
  year={2023}
}

@article{friedrich2023fair,
  title={Fair diffusion: Instructing text-to-image generation models on fairness},
  author={Friedrich, Felix and Brack, Manuel and Struppek, Lukas and Hintersdorf, Dominik and Schramowski, Patrick and Luccioni, Sasha and Kersting, Kristian},
  journal={arXiv preprint arXiv:2302.10893},
  year={2023}
}

@article{seshadri2023bias,
  title={The bias amplification paradox in text-to-image generation},
  author={Seshadri, Preethi and Singh, Sameer and Elazar, Yanai},
  journal={arXiv preprint arXiv:2308.00755},
  year={2023}
}

@article{su2023manifold,
  title={Manifold-guided sampling in diffusion models for unbiased image generation},
  author={Su, Xingzhe and Qiang, Wenwen and Song, Zeen and Gao, Hang and Wu, Fengge and Zheng, Changwen},
  journal={arXiv preprint arXiv:2307.08199},
  volume={9},
  year={2023}
}

@article{liu2023vida,
  title={Vida: Homeostatic visual domain adapter for continual test time adaptation},
  author={Liu, Jiaming and Yang, Senqiao and Jia, Peidong and Zhang, Renrui and Lu, Ming and Guo, Yandong and Xue, Wei and Zhang, Shanghang},
  journal={arXiv preprint arXiv:2306.04344},
  year={2023}
}

@article{zheng2026pearl,
  title={PEARL: Personalized Streaming Video Understanding Model},
  author={Zheng, Yuanhong and An, Ruichuan and Lin, Xiaopeng and Liu, Yuxing and Yang, Sihan and Zhang, Huanyu and Li, Haodong and Zhang, Qintong and Zhang, Renrui and Li, Guopeng and others},
  journal={arXiv preprint arXiv:2603.20422},
  year={2026}
}

@article{feng2026m2a,
  title={M2a: Multimodal memory agent with dual-layer hybrid memory for long-term personalized interactions},
  author={Feng, Junyu and Xu, Binxiao and Chen, Jiayi and Dai, Mengyu and Wu, Cenyang and Li, Haodong and Zeng, Bohan and Xie, Yunliu and Liang, Hao and Lu, Ming and others},
  journal={arXiv preprint arXiv:2602.07624},
  year={2026}
}

@article{xu2025jarvis,
  title={Jarvis: Towards Personalized AI Assistant via Personal KV-Cache Retrieval},
  author={Xu, Binxiao and Feng, Junyu and Lu, Shaolin and Luo, Yulin and Yan, Shilin and Liang, Hao and Lu, Ming and Zhang, Wentao},
  journal={arXiv preprint arXiv:2510.22765},
  year={2025}
}

@article{yang2025small,
  title={Small-Large Collaboration: Training-efficient Concept Personalization for Large VLM using a Meta Personalized Small VLM},
  author={Yang, Sihan and Ji, Huitong and Lu, Shaolin and Chen, Jiayi and Xu, Binxiao and Lu, Ming and Zhang, Yuanxing and Dong, Wenhui and Zhang, Wentao},
  journal={arXiv preprint arXiv:2508.07260},
  year={2025}
}
